\documentclass{article}

\PassOptionsToPackage{square,numbers}{natbib}

    \usepackage[preprint]{neurips_2021}

\usepackage[utf8]{inputenc} 
\usepackage[T1]{fontenc}    
\usepackage{hyperref}       
\usepackage{url}            
\usepackage{booktabs}       
\usepackage{amsfonts}       
\usepackage{nicefrac}       
\usepackage{microtype}      
\usepackage{xcolor} 
\usepackage{soul,xcolor}
\setstcolor{blue}
\usepackage{subfigure}
\usepackage{multirow}
\usepackage{bm}
\usepackage{stackengine}
\usepackage{tabularx,makecell}
\usepackage{wrapfig}
\usepackage[pdftex]{graphicx}
\newcommand*\diff{\mathop{}\!\mathrm{d}}  
\usepackage{amsmath,amsthm,amsfonts,amssymb,mathtools}
\usepackage{algpseudocode,algorithm,algorithmicx}
\algrenewcommand\algorithmicrequire{\textbf{Input:}}
\algrenewcommand\algorithmicensure{\textbf{Output:}}
\algnewcommand\algorithmicinput{\textbf{Input:}}
\algnewcommand\algorithmicoutput{\textbf{Output:}}
\algnewcommand\Input{\item[\algorithmicinput]}%
\algnewcommand\Output{\item[\algorithmicoutput]}%
\usepackage{commath}
\usepackage{caption}
\usepackage{xcolor}
\usepackage{float}
\usepackage{floatflt}
\definecolor{xblue}{RGB}{124,175,211}
\definecolor{xgreen}{RGB}{127,191,127}
\definecolor{xblue2}{RGB}{64,64,255}
\definecolor{xgreen2}{RGB}{175,215,175}
\definecolor{xorange2}{RGB}{255,210,127}
\newtheorem{theorem}{Theorem}[section]

\newtheorem{lemma}[theorem]{Lemma}

\newcommand\T{\rule{0pt}{2.6ex}}       
\newcommand\B{\rule[-1.2ex]{0pt}{0pt}} 

\title{Incorporating NODE with Pre-trained Neural Differential Operator for Learning Dynamics}

\author{%
  Shiqi Gong$^{1,3}$\thanks{This work was done when the author was visiting Microsoft Research Asia.}, Qi Meng$^2$, Yue Wang$^2$, Lijun Wu$^2$, Wei Chen$^2$\\
  \textbf{Zhi-Ming Ma}$^{1,3}$, \textbf{Tie-Yan Liu}$^2$  \\
  $^1$University of Chinese Academy of Sciences, $^2$Microsoft Research Asia, \\ $^3$Academy of Mathematics and Systems Science, CAS \\
  $^1$\texttt{gongshiqi15@mails.ucas.ac.cn}, $^3$\texttt{mazm@amt.ac.cn}, \\
   $^2$\texttt{\{meq, yuwang5, lijuwu, wche, tyliu\}@microsoft.com}\\
}

\begin{document}

\maketitle

\begin{abstract}
Learning dynamics governed by differential equations is crucial for predicting and controlling the systems in science and engineering. 
Neural Ordinary Differential Equation (NODE), a deep learning model integrated with differential equations, is popular in learning dynamics recently due to its robustness to irregular samples and its flexibility to high-dimensional input. However, the training of NODE is sensitive to the precision of the numerical solver, which makes the convergence of NODE unstable, especially for ill-conditioned dynamical systems. 
In this paper, to reduce the reliance on the numerical solver, we propose to enhance the supervised signal in the training of NODE. 
Specifically, we pre-train a neural differential operator (NDO) to output an estimation of the derivatives to serve as an additional supervised signal.  The NDO is pre-trained on a class of basis functions and learns the mapping between the trajectory samples of these functions to their derivatives. 
 To leverage both the trajectory signal and the estimated derivatives from NDO, we propose an algorithm called NDO-NODE, in which the loss function contains two terms: the fitness on the true trajectory samples and the fitness on the estimated derivatives that are outputted by the pre-trained NDO. Experiments on various kinds of dynamics show that our proposed NDO-NODE can consistently improve the forecasting accuracy with one pre-trained NDO. Especially for the stiff ODEs, we observe that NDO-NODE can capture the transitions in the dynamics more accurately compared with other regularization methods. 
\end{abstract}
\raggedbottom

  
\section{Introduction}

    Learning dynamics governed by differential equations is crucial for predicting and controlling the systems in science and engineering such as predicting future movements of planets in physics, protein structure prediction \cite{zhang2008progress}, the evolution of fluid flow \cite{wiewel2019latent} and many other applications \cite{small2005applied}. The recently proposed neural ordinary differential equations (Neural ODEs) \cite{chenNeuralOrdinaryDifferential2018}, a deep learning model integrated with differential equations, shows great promise in the scientific field \cite{guenAugmentingPhysicalModels2020,norcliffeSecondOrderBehaviour2020,zhuangAdaptiveCheckpointAdjoint2020,lee2020parameterized,chenLearningNeuralEvent2020}. The continuous nature of NODEs and their differential equation structure of the hypothesis have made them particularly suitable for learning the dynamics of complex physical systems. Its robustness to irregular samples and flexibility to high-dimensional input make it superior compared with traditional non-deep learning based dynamics identification methods \cite{aliee2021beyond, zhong2021benchmarking,duong2021hamiltonian}.
    
     NODE models the higher-order derivatives directly from the discrete trajectory samples (i.e., the coordinates of the object as a series of discrete-time points).  This process makes the training of NODE highly rely on the numerical ODE solver.
     

\begin{wrapfigure}{r}{5.5cm}
    \vspace*{0.2cm}
        \centering
        \includegraphics[width=5.5cm]{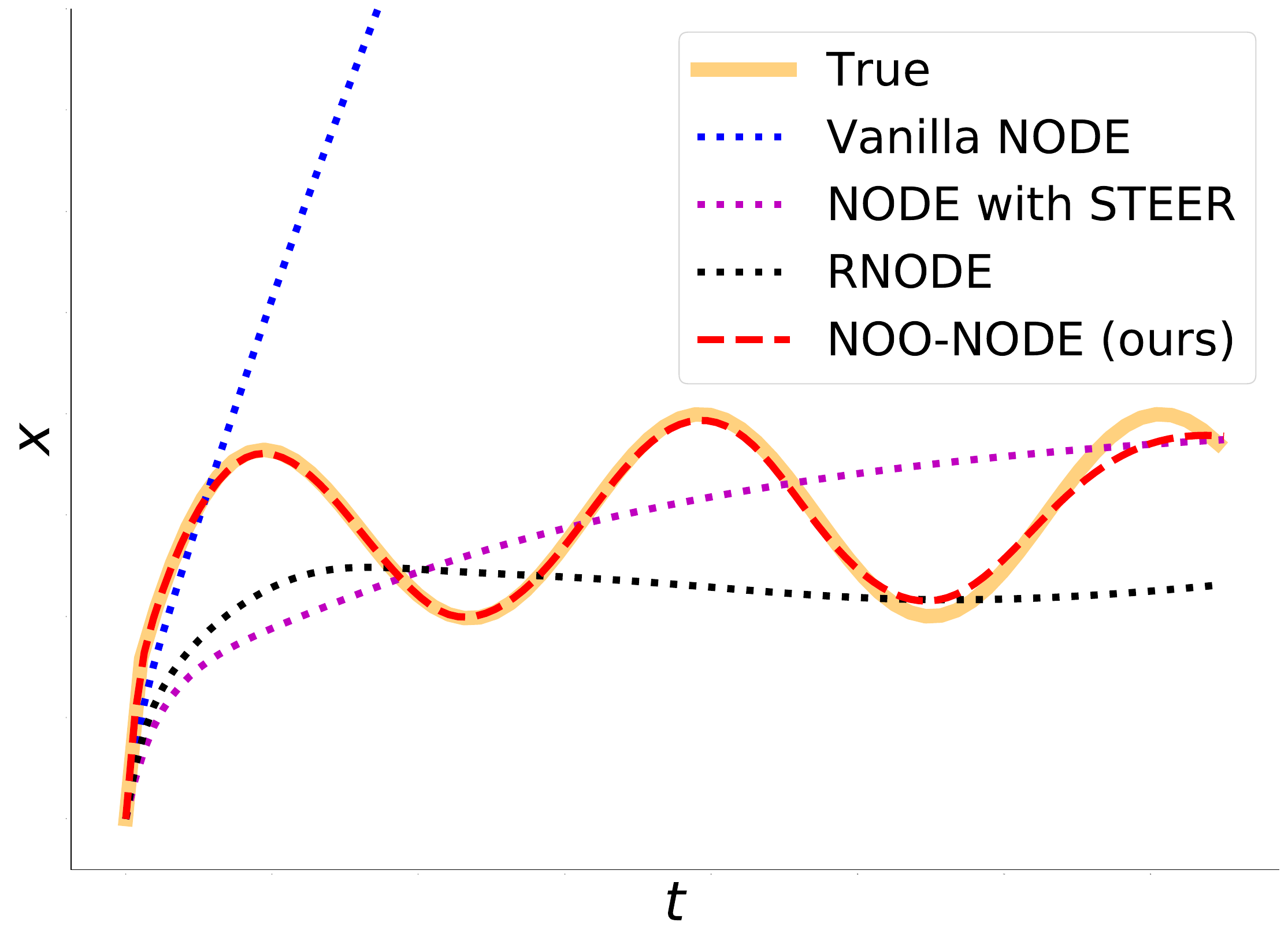}
        \caption{NDO-NODE learns the stiff ODE $\frac{dx}{dt} = -1000x+3000-2000e^{-t} + 1000\sin(t)$ well, while others fail.} 
        \label{fig:corner}
    \vspace{-0.4cm}
\end{wrapfigure}
 
Recent works \cite{ijcai2019-103,zhuangAdaptiveCheckpointAdjoint2020} show that 
the solver can amplify numerical noise and be numerically unstable, even lead to incorrect derivative signal and divergence of the training, especially for ill-conditioned dynamical systems such as stiff ODEs or chaotic systems \cite{kim2021stiff,ghosh2020steer}.
There are some regularization methods designed for NODE to force the learned dynamics to be simple and avoid the instability such as randomizing end temporal point \cite{ghosh2020steer} and regularizing high-order derivatives \cite{kelly2020learning,finlayHowTrainYour2020a}. These techniques mainly focus on simplifying or smoothing the learned model but they may miss the important transition point of the dynamics as illustrated in Figure~\ref{fig:corner}.

In this paper, we propose another approach to stabilize the training of NODE, which 
leverages the knowledge on derivative calculation mined from various types of functions to guide the training of NODE.
The intuition is that: the differential operator is common for different functions and the learned knowledge can be transferred to the target dynamics (which has been verified in other machine learning tasks \cite{yang2019xlnet,devlin2018bert,radford2018improving,He_2019_ICCV,goyal2017accurate,yanai2015food}).
 We mine the knowledge by pre-training a neural differential operator (NDO) on a pre-designed library of basis functions. Specifically, 
the basis functions are composed of diverse types of functional basis such as triangle basis and polynomial basis. The NDO maps the trajectories of these functions to their derivatives.  
We test on various dynamics and show that the estimations of NDO are more accurate and robust compared with traditional derivative estimation methods such as finite difference.

Since the pre-training process does not rely on the numerical solver, we leverage the output of NDO as an auxiliary supervised signal to help the training of NODE. Specifically, when we learn unknown dynamics, we first input the training points of the dynamics to the pre-trained NDO to get the estimated derivatives. 
Then we use these estimations as another supervised signal in the training process of NODE by constraining the distance between estimated derivatives and the output derivatives of NODE. We name this neural ODEs algorithm as \emph{NDO-NODE}.

We conduct experiments on various dynamics including physical dynamics, stiff ODEs, and real-world dynamics governed by differential equations to verify the effectiveness of NDO-NODE. Experiments show that NDO-NODE can consistently improve both the interpolation and extrapolation accuracy on these tasks with one pre-trained NDO. Furthermore, we observe that NDO-NODE is robust to 
noisy and irregular-time observations compared with the baselines.



\begin{figure*}[t]
  \centering
  \includegraphics[width=\textwidth]{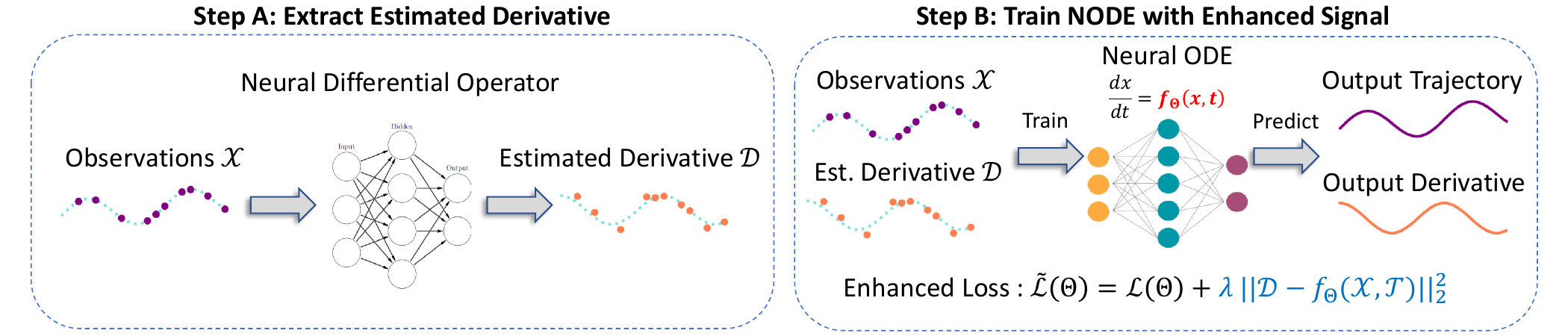}
  \caption{Illustrations of the proposed algorithm NDO-NODE. \textbf{(A)} Extract estimated derivative from trajectory observations by a pre-trained neural differential operator. \textbf{(B)} Incorporate auxiliary derivative signal into the training of NODE. }
  \label{fig:workflow}
\end{figure*}

\section{Background}
\subsection{Neural ODEs}
Neural ODEs \cite{chenNeuralOrdinaryDifferential2018} are a new family of deep neural network models that parameterize the derivative of the continuous state using a neural network. This model can be represented by ODEs:
{\small\begin{align}\label{eq:ode}
    \frac{\diff x(t)}{\diff t} = f_\Theta (x,t), \quad \text{   s.t.  } x{(t_0)} = x_0,
\end{align}}

where $x(t)$ is the vector of continuous state, $f_\Theta$ is any kind of network with parameters $\Theta$, and $x_0$ is the input state at the initial time $t_0$. State $x{(t_i)}$ can be computed by solving the initial value problem (IVP): $x{(t_i)} = x_0 + \int_{t_0}^{t_i} f_\Theta (x,t) \diff t = \text{ODESolve}(x_0,f_\Theta,t_0,t_i)$, which can be done by any numerical ODE solver.

Suppose we want to learn unknown dynamics by a sequence of trajectory observations $\mathcal{X} = (x_0,x_1,\cdots,x_N)$ at times $\mathcal{T}=(t_0,t_1,\cdots,t_N)$. The training process of NODE can be formulated by following optimization problem:
{\small\begin{align}\label{eq:loss0}
    \min _\Theta \mathcal{L}(\mathcal{X}',\mathcal{X}) &= \frac{1}{N} \sum_{i=0}^N \ell(x'_i,x_i) \\
    \text{s.t. } \mathcal{X}' = (x'_0,x'_1,\cdots,x'_N)&= \text{ODESolve}(x_0,f_\Theta,t_0,\mathcal{T}) ~,
\end{align}}
where $\mathcal{X}'$ is the predictions and $\ell(\cdot,\cdot)$ is a distance metric, e.g., $L_1$, $L_2$ distance.

In this paper, we will study NODE on several types of ODE systems including stiff ODEs and chaotic ODEs. Stiff ODEs, which very slow and very fast transients coexist in their solution \cite{thohura2013numerical}, are frequently raised from chemical kinetic systems \cite{shieh1988evaluation} and biological systems \cite{haefner2005modeling}. The stiffness may cause numerically unstable unless an extremely small step size is used. Thus neural ODEs need to take more time to learn stiff systems and even fail to learn in many cases \cite{kim2021stiff}. Chaotic ODEs exist in modeling many natural systems such as fluid flow \cite{lorenz1963deterministic}, weather and climate \cite{ivancevic2008complex}. They are extremely sensitive to infinitesimal perturbations in initial conditions, which makes prediction difficult.


\subsection{Related Works}
\subsubsection{Neural ODEs and Its Variants}
\citet{chenNeuralOrdinaryDifferential2018} proposed the first neural ODEs model and it has been a popular model in scientific fields due to the continuous property. {Several recent works adjust the structures of NODE to be more suitable for some specific kinds of ODEs such as second-order ODEs \cite{norcliffeSecondOrderBehaviour2020}, piecewise ODEs \cite{chenLearningNeuralEvent2020,greydanus2021piecewise}, stochastic differential equations (SDEs) \cite{jiaNeuralJumpStochastic2020}, controlled differential equations (CDEs) \cite{kidgerNeuralControlledDifferential2020,morrill2021neural} and parameterized neural ODEs \cite{lee2020parameterized}.} 
\citet{guenAugmentingPhysicalModels2020} merge data-driven neural ODEs with physical prior knowledge to augment incomplete physical dynamics. 

To stablize and accelerate the training of NODE, some regularization methods are proposed for NODE. \citet{finlayHowTrainYour2020a} and \citet{kelly2020learning} propose to regularize the $L_2$ norm to encourage NODE to learn simpler dynamics. 
\citet{ghosh2020steer} propose to randomly sample the end time of the ODE during training. 
Our proposed method is vertical to these works and we leverage the knowledge on basis functions to serve as an auxiliary signal in NODE training.

\subsubsection{Other Dynamic Identification Methods}
Besides neural ODEs which directly model the derivatives, \citet{greydanusHamiltonianNeuralNetworks2019a,cranmerLagrangianNeuralNetworks2020a} model the Langrangian or Hamiltonian of the dynamics to keep the symmetry or conservation law in physics. However, these methods require prior knowledge about the systems and may need to design and adjust case by case. There are also some traditional methods to learn dynamical systems such as state observers in control theory \cite{niethammer2001parameter,levant2017sliding,bhasin2012robust}. These methods assume specific structure of the dynamics \cite{bhasin2012robust} or rely on regularly discretized samples \cite{levant2017sliding}, while NODE is applicable to approximate any continuous dynamics with irregular samples and less prior knowledge on its structure.

\section{Algorithm: NDO-NODE}
\label{sec:alg}
In this section, we first introduce the NDO-NODE algorithm which learns the model in Equation~(\ref{eq:ode}) by incorporating estimated derivatives. Then, we introduce the learning of neural differential operator to obtain the estimated derivatives.

\subsection{NDO-NODE Framework}
\begin{figure}[h]
    \centering
\begin{minipage}{0.85\linewidth}
\begin{algorithm}[H]
    \caption{NDO-NODE}
    \label{alg:ndonode}
    \begin{algorithmic}[1]
        \Input Trajectory observation vector $\mathcal{X}$ at times $\mathcal{T}$, query time vector  $\widetilde{\mathcal{T}}$, neural differential operator $\text{NDO}(\cdot)$, derivative signal strength $\lambda$
        \Output Prediction vector $\widetilde{\mathcal{X}}$ at times $\widetilde{\mathcal{T}}$
        \State Initialize $f_{\Theta}$ in neural ODEs $\frac{\diff x_t}{\diff t} = f_{\Theta}(t,x_t)$ 
        \State Estimate derivative $\mathcal{D} = \text{NDO}(\mathcal{X},\mathcal{T})$ 
        \Repeat
            \State $\mathcal{X}'$ = ODESolve($x_0$,$f_{\Theta}$,$t_0$,$\mathcal{T}$)
            \State $\mathcal{\widetilde{L}} =  \mathcal{L}(\mathcal{X}',\mathcal{X}) + \lambda \cdot \|\mathcal{D} - f_\Theta (\mathcal{X},\mathcal{T}) \|_2^2$
            \State Update parameters $\Theta$ by $\nabla_\Theta \widetilde{\mathcal{L}}$
        \Until {converge}
        \State \Return $\widetilde{\mathcal{X}}=$ ODESolve($x_0$,$f_{\Theta}$,$t_0$,$\widetilde{\mathcal{T}}$)
    \end{algorithmic}
\end{algorithm}
\end{minipage}
\end{figure}

As mentioned previously, to enhance the supervised signal of NODE in learning dynamics, we propose to incorporate an estimation on the derivatives into the learning process, which directly constrains the underlying derivatives of NODE.  The new loss function can be expressed as
\begin{align}
    \mathcal{\widetilde{L}} = \mathcal{L}(\mathcal{X}',\mathcal{X}) + \lambda \cdot\|\mathcal{D} - f_\Theta (\mathcal{X},\mathcal{T}) \|_2^2 ,
\end{align}
where $\mathcal{X}' = (x'_0,x'_1,\cdots,x'_N) = \text{ODESolve}(x_0,f_\Theta,t_0,\mathcal{T})$ denotes the predictions of NODE,  $\mathcal{D} = (d_0, \cdots,d_N)$ denotes the corresponding underlying derivatives at time points $\mathcal{T}$, and $\lambda$ controls the strength of the derivative signal. The detailed NDO-NODE algorithm is shown in Algorithm \ref{alg:ndonode}. 

Note that, although we do not know the ground truth derivative $\mathcal{D}$, the principle of obtaining the derivatives for continuous functions is common. In the next section, we introduce how to pre-train a  neural differential operator ($\text{NDO}(\cdot)$) to estimate the derivative $\mathcal{D}$ from the discrete trajectory samples. As shown in Algorithm \ref{alg:ndonode}, we can leverage NDO to generate the estimated derivative $\mathcal{D}$ firstly (Line 2) and then use it as an extra term into the loss function (Line 5).

\subsection{Neural Differential Operator}
In this section, we introduce the details to train the neural differential operator. 
\begin{figure}[h]
    \centering
\begin{minipage}{0.85\linewidth}
\begin{algorithm}[H]
	\caption{Learning Neural Differential Operator $\text{NDO}(\cdot)$ via Library $\mathcal{Z}_{lib}$}
	\label{alg:NDO}
	\begin{algorithmic}[1]
		\Input Function library $\mathcal{Z}_{lib}$
		\Output Neural differential operator $\text{NDO}(\cdot)$
		\State Initialize network $\text{NDO}(\cdot)$
		\Repeat
		\State Randomly draw function $z \in \mathcal{Z}_{lib}$
		\State Generate discretized times: $\mathcal{T}_z = (t_0,t_1,\cdots,t_N)$ 
		\State $\mathcal{X}_z = \left(z(t_0),z(t_1),\cdots,z(t_N)\right)$
		\State $\dot{\mathcal{X}}_z = (\dot{z}(t_0),\dot{z}(t_1),\cdots,\dot{z}(t_N))$
		\State  $\mathcal{L} = \|\text{NDO}(\mathcal{X}_z,\mathcal{T}_z) - \dot{\mathcal{X}}_z\|_2^2$
		\State Update $\text{NDO}(\cdot)$ by $\nabla \mathcal{L}$
		\Until {converge}
	\end{algorithmic}
\end{algorithm}
\end{minipage}
\end{figure}

 A neural differential operator $\text{NDO}(\cdot):\mathbb{R}^N\times [0,T] \rightarrow \mathbb{R}^N, (\mathcal{X},\mathcal{T}) \mapsto \mathcal{D}$, is a sequence to sequence model that projects the trajectory vector $\mathcal{X}$ at times $\mathcal{T}$ to corresponding estimated derivative $\mathcal{D}$.  Generally speaking, the neural differential operator does not depend on any specific network architecture. Networks such as fully connected neural networks and CNN can be used for trajectories with fixed time grid, while sequence model like LSTM, RNN, and transformers \cite{NIPS2017_3f5ee243} can cope with irregular times better.

To train our NDO, we need trajectory samples $(\mathcal{X},\mathcal{T})$ together with their true derivatives $\dot{\mathcal{X}}$ as labels. We propose to generate the training data from a synthetic function library since it will provide us enough data without additional work, such as conducting physical experiments or manually labeling. 
Because any continuous function can be expanded to polynomial series and trigonometric series \cite{achieser2013theory}, we use polynomial and trigonometric functions as basis functions to construct the library in our work. Now a library can be written as the linear combination of bases: 
{\tiny\begin{align}
\mathcal{Z}_{lib} = 
\left\{\sum_{i=0}^P \left[ a_i \sin(it) + b_i \cos(it) \right] + \sum_{i=0}^Q c_i t^i \middle| P,Q \in \mathbb{N}, \abs{a_i},\abs{b_i},\abs{c_i}< C  \right\}  \label{eq:lib}, 
\end{align}}
where $a_i,b_i,c_i$ are coefficients of bases, hyperparameters $P,Q$ control the complexity of a library, and $C$ controls the scale of a library. Given a library, our training data can be generated from the discretization values of random function sample $z \in \mathcal{Z}_{lib}$ and corresponding derivative $\dot{z}$. Specifically, for each training data, first we uniformly draw some bases with coefficients from $\mathcal{Z}_{lib}$, and add them up to get the random function $z$. Its derivative $\dot{z}$ can be computed symbolically. Next, we uniformly sample time points $\mathcal{T}_z=(t_0,t_1,\cdots,t_N)$ from the fixed time interval $[T_0,T_1]$. \footnote{In experiments, without loss of generality, we fix $[T_0,T_1]$ to be $[0,1]$ as any time interval can be rescaled to it.} Finally, the evaluations $\mathcal{X}_z= \left(z(t_0),z(t_1),\cdots,z(t_N)\right)$ and $\dot{\mathcal{X}}_z=(\dot{z}(t_0),\dot{z}(t_1),\cdots,\dot{z}(t_N))$ at times $\mathcal{T}_z$ are the training inputs and labels, respectively. After generating training data, we minimize $L_2$ loss between the output of $\text{NDO}(\cdot)$ and labels $\dot{\mathcal{X}}_z$. To sum up, the training process for neural differential operator is shown in Algorithm \ref{alg:NDO}. Note that, NDO is compatible with all orders of derivatives. For $k$-th order NDO, we just need to simply change the label to $k$-th order derivatives and input to $(k-1)$-th order derivatives. 


 Next, we provide the theoretical guarantee on learning the differential operator by neural networks. It has been shown that neural networks are universal approximators for non-linear operators in \citep{chen1995universal,lu2021learning}.  In the next proposition, we show that the error for the learned neural differential operator $\text{NDO}(\cdot)$ and the ground truth derivative for a given continuous differentiable function $g(t):[0,1]\rightarrow\mathbb{R}$.
 
 \begin{theorem}\label{thm1}Suppose that $\mathcal{Z}_{lib}'\subset\mathcal{Z}_{lib}$ is the training function set for NDO. The Lipschitz constant for the learned neural differential operator function is $L_{NN}$. For a given continuous differentiable function $h(t):[T_0,T_1]\rightarrow\mathbb{R}$, we define the distance between two functions as $\rho(h,z)=\sum_{i=1}^N|h(t_i)-z(t_i)|$, where $\{t_i\}_{i=1}^N$ equally partition the time interval $[T_0,T_1]$. $z(t)\in\mathcal{Z}'_{lib}$ is a function in the training data, $h(t)$ is an arbitrary function. The output derivative of NDO for a function is denoted by the subscription $(\cdot)_{\text{NDO}}$. Then the error of the output derivation $\dot{h}_{\text{NDO}}$ and the ground truth derivative $\dot{h}$ can be upper bounded as:
{\small\begin{align*}
\rho({\dot{h}}_{\text{NDO}},\dot{h})
\le &\min_{z(t) \in \mathcal{Z}'_{lib}} \left\{ L_{NN}\int_{T_0}^{T_1}|z(t)-h(t)|dt   +\int_{T_0}^{T_1}|\dot{z}(t)-\dot{h}(t)|dt 
+\frac{|T_1-T_0|^3}{12N^2}M + \rho({\dot{z}_{\text{NDO}}},\dot{z}) \right\},
\end{align*}where $M=L_{NN}\cdot\max_{t\in[T_0,T_1]}|\ddot{\epsilon}(t)|+\max_{t\in[T_0,T_1]}|\dddot{\epsilon}(t)|$ with $\epsilon(t)=|z(t)-h(t)|$.}
 \end{theorem}

Theorem \ref{thm1} shows that the upper bound of $\rho({\dot{h}}_{\text{NDO}},\dot{h})$ depends on three factors: the approximation error of $z(t)$ (measured by both the $L_1$ distance between $z(t)$ and $h(t)$ and the $L_1$ distance between $\dot{z}(t)$ and $\dot{h}(t)$), the smoothness of the NN model $L_{NN}$ and the optimization error on the training data $z$, i.e., $\rho(\dot{z}_{\text{NDO}},\dot{z})$. As the library becomes large, the approximation error has chances to become small but the optimization will become hard (which may cause the increase of $\rho(\dot{z}_{\text{NDO}},\dot{z})$). Because our library is constructed by the basis functions that can universally approximate any continuous functions, it can well approximate the ground truth derivatives with more bases. Moreover, Theorem \ref{thm1} can be easily extended to $h(t)\in\mathbb{R}^d$, because all the $d$ outputs are functions of time $t$. More details can be found in Appendix. 
 
\subsection{Empirical Studies on NDO}\label{sec:ndo}
\subsubsection{NDO Settings}
We use an LSTM model to implement $\text{NDO}(\cdot)$.  
We set $t_0=0$ and $t_N=1$ for the inputs to make the training simpler. \footnote{When we use pre-trained NDO for downstream tasks, we can always standardize the time point in $\mathcal{T}$ by multiplying a factor $1/t_{N}$ to both $\mathcal{T}$ and output of NDO if $t_N \neq 1$.}
We augment the input sequence as $(\mathcal{X},\mathcal{T},\Delta\mathcal{T}) = \left\{(x_i,t_i,\Delta t_i)\right\}_{i=0}^N$, where $\Delta t_i = t_i-t_{i-1}$, to serve for this task better.
For fixed hyperparameters $(P,Q,C)$ of library $\mathcal{Z}_{lib}$, we randomly draw $10000$ functions from $\mathcal{Z}_{lib}$ and discretize them by $100$ uniformly random times in the interval $[0,1]$ as our training data. More training details can be found in Appendix.

\subsubsection{Estimation Accuracy of NDO vs. Library Complexity}
  \begin{figure*}[t!]
 	\centering
 	\subfigure[$P=0$
 	]{\includegraphics[width=0.19\textwidth]{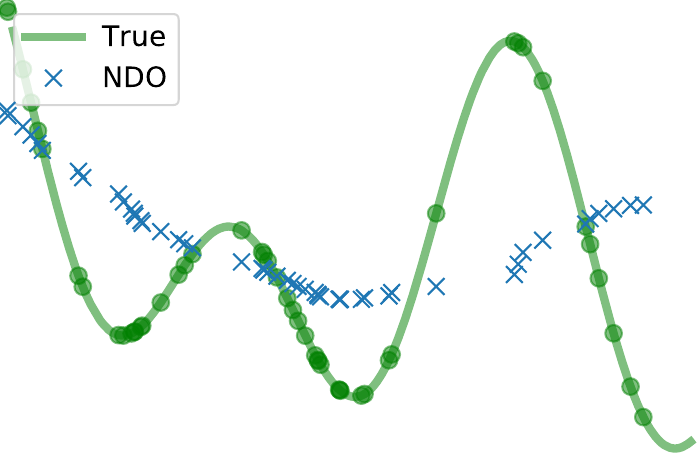}}
	\subfigure[$P=5$
	]{\includegraphics[width=0.19\textwidth]{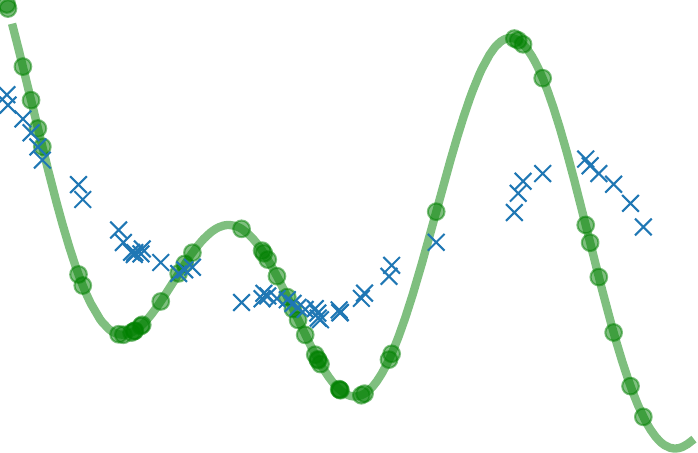}}
	\subfigure[$P=20$
	]{\includegraphics[width=0.19\textwidth]{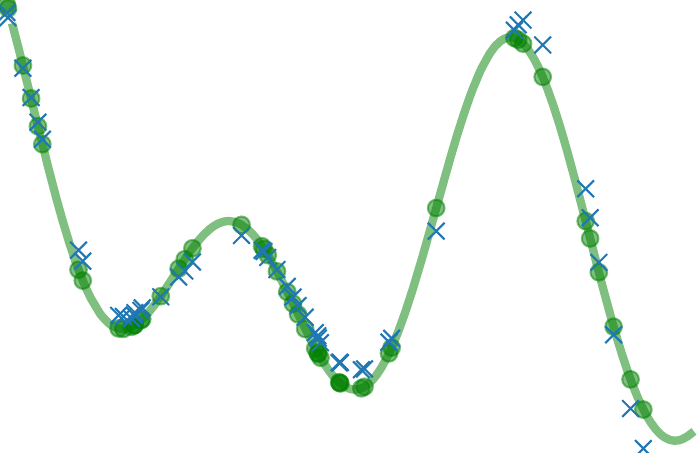}}
	\subfigure[$P=50$
	]{\includegraphics[width=0.19\textwidth]{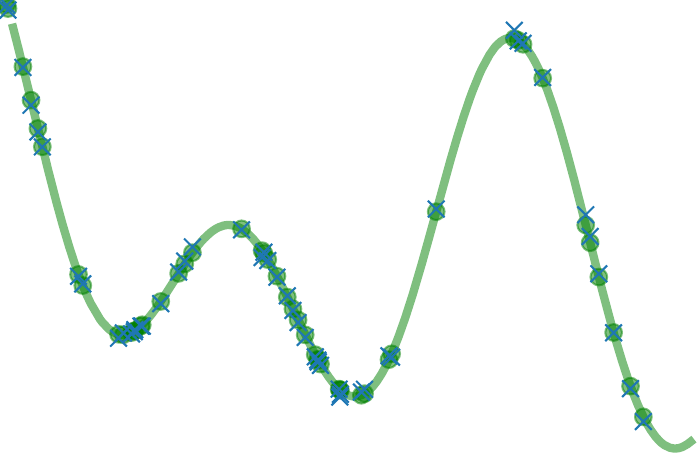}\label{fig:grad_ndo}}
 	\subfigure[\stackanchor{Finite Difference}{Five-point Stencil}]{\includegraphics[width=0.19\textwidth]{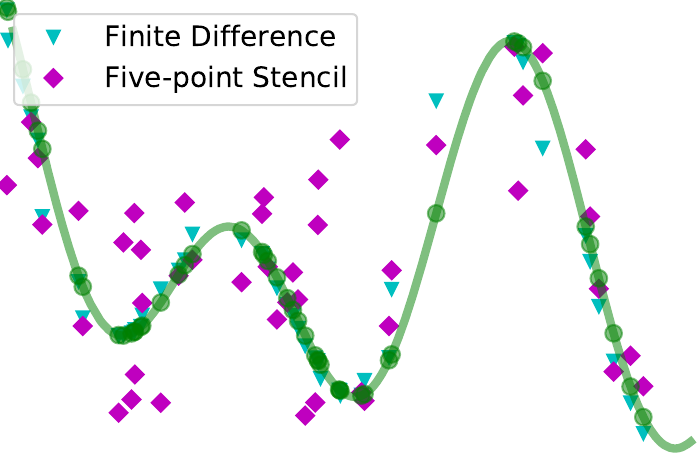}\label{fig:grad_finite}} 
     \caption{Estimate the derivative of {\small$z(t) = \sin(3t)/3+ \sin(15t)/15+\sin(30t)/30,~ t \in [0,1]$}. \textbf{(a-d)}: Results for NDO pre-trained by libraries with different parameter $P$. \textbf{(e)}: Results for finite difference and five-point stencil method. }
 	\label{fig:diver}
 \end{figure*}
 
To test how the library complexity affects the accuracy of the estimation, we pre-train NDOs with different libraries and use them on a specified $z(t)$ (see Figure~\ref{fig:diver}). We fix $(Q,C)=(3,10)$ and set different values for $P$ \footnote{We only demonstrate the results for different values for $P$ here and we put more results for $Q$ in Appendix.}. Note that $z(t)$ has 3 terms that are successively contained into the libraries with $P=5,20,50$. From Figure~\ref{fig:diver}, we observe that the accuracy of estimations becomes higher when more terms get included in the library. When the library can not cover $z(t)$, i.e., $P=0,5,20$, NDO will fit the main parts that the library contains. When the library fully covers $z(t)$, i.e., $P=50$, the estimations get very close to the ground truth derivatives, which agrees with our theoretical results. Therefore, for the experimental results shown in the following subsections, we use the NDO trained with  $(P,Q,C)=(50, 3, 10)$.

\subsubsection{Accuracy of Estimated Derivatives}

\begin{figure*}[h]
 	\centering
 	\subfigure[Planar spiral systems ]{\includegraphics[width=0.24\textwidth]{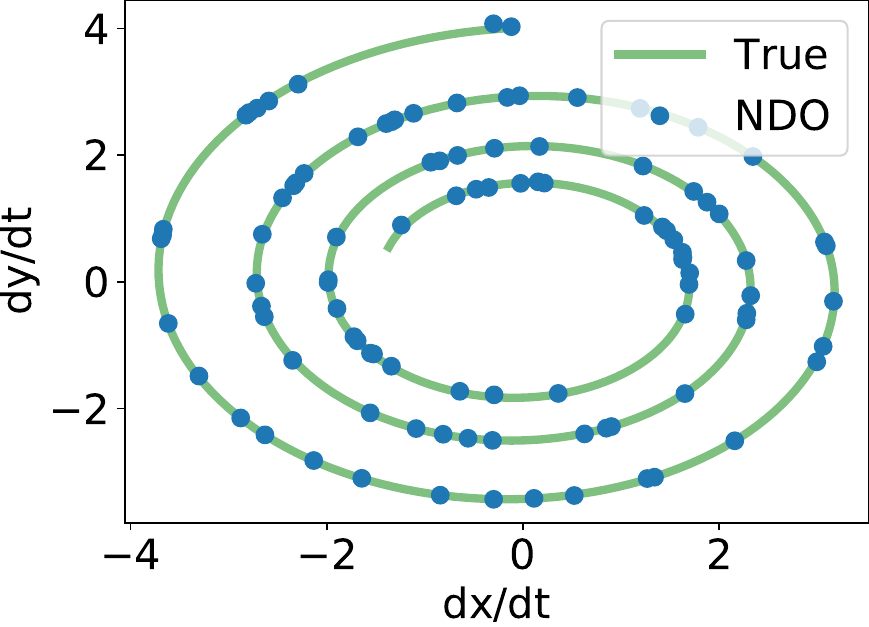}} 
	\subfigure[Damped harmonic oscillator]{\includegraphics[width=0.24\textwidth]{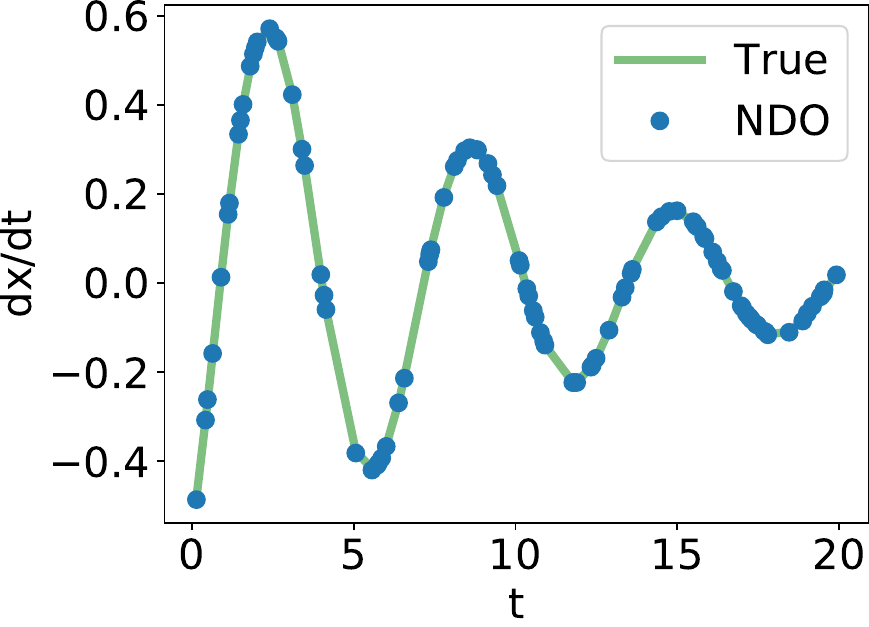}}
	\subfigure[Stiff ODE in Eq.(9)]{\includegraphics[width=0.23\textwidth]{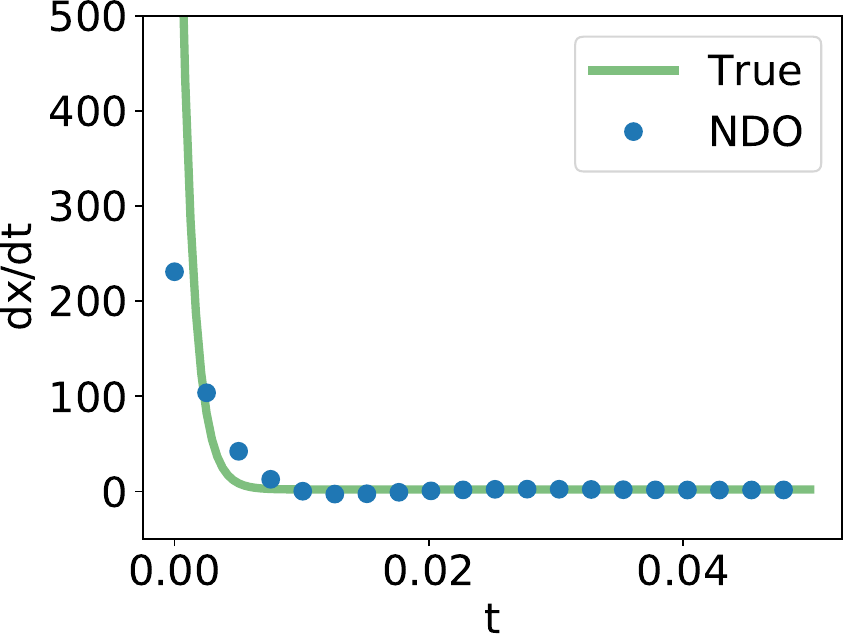}}
	\subfigure[One traj in three-body problem]{\includegraphics[width=0.24\textwidth]{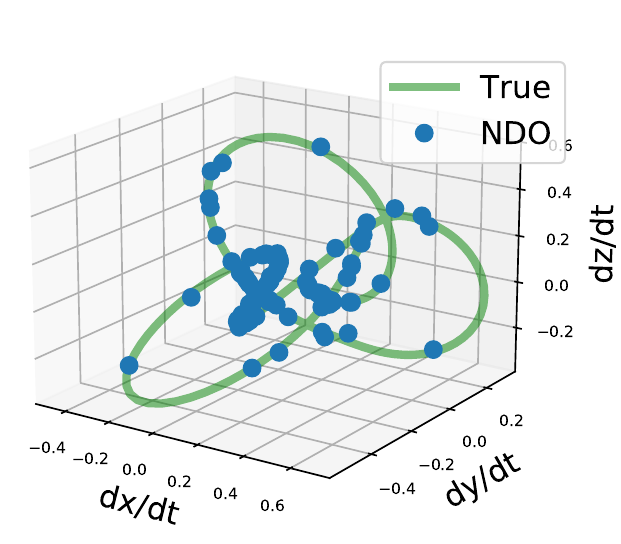}}
 	\caption{Derivative comparisons between NDO estimations and the ground truth derivative.} 
	\label{fig:grad_comp}
\end{figure*}

We compare NDO model with $(P,Q,C)=(50, 3, 10)$ to traditional derivative estimation methods including finite difference and five-point stencil \cite{10.5555/2161609}.
As shown in Figure~\ref{fig:grad_finite}(d) and Figure~\ref{fig:grad_ndo}, the results for finite difference and five-point stencil perform worse than NDO due to the irregular time-series data. 
We also apply our pre-trained NDO model with $(P,Q,C)=(50, 3, 10)$ to dynamics with different characters such as chaotic, stiff, and periodically oscillated \footnote{The details for the dynamics will be introduced in the next section.}.  We compare the output of NDO with the ground truth. From Figure~\ref{fig:grad_comp}, we observe that: 1) the output of NDO matches the ground truth derivatives well on periodically oscillated dynamics and chaotic dynamics; 2) for stiff ODEs, the estimations of NDO match the overall curvature of the ground truth except that it is smoother than ground truth at the sharp point. It is because the NDO is learned on a set of functions in which the smoothness are controlled. 

\begin{table*}[h]
\centering
\caption{Mean squared error (mean $\pm$ std, $\times 10^{p}$, over 3 runs) for  experiments on physical systems. For planar spiral systems, damped  harmonic  oscillator, and three-body problem, In. MSEs (interpolation MSE) are computed on range $[0,5],[0,10],[0,1]$, Ex. MSEs (extrapolation MSE) are computed on range $[5,10],[10,20],[1,2]$, and scale factor $p=-2,-3,-1$, respectively.  }
\label{table:threebody}
\begin{tabular}{ccccc}
\toprule
System & MSE & NODE & RNODE & NDO-NODE(ours) \\
\midrule 
\multirow{2}{*}{Planar Spiral Systems} & In.   & $0.026 \pm 0.014$     & $0.040 \pm 0.019$     & $\bm{0.012 \pm 0.004}$                        \B \\ \cline{2-5} 
                         & Ex.     & $4.524 \pm 2.890$     & $4.023 \pm 2.749$     & $\bm{0.525 \pm 0.304}$    \T                \\  
\midrule
\multirow{2}{*}{Damped  Harmonic  Oscillator} & In.            & $5.217 \pm 1.948$       & $2.058 \pm 0.872$       &$\bm{ 1.005 \pm 0.305}$           \B \\\cline{2-5} 
                         & Ex.              & $7.918 \pm 5.762$       & $4.239 \pm 2.058$       & $\bm{ 0.860 \pm 0.336}$       \T      \\  
\midrule
\multirow{2}{*}{Three-body Problem} & In.  & $0.414 \pm 0.110$       & $0.393 \pm 0.040$ & $\bm{0.274 \pm                           0.063}$        \B          \\ \cline{2-5} 
                         & Ex.      & $6.908 \pm 2.077$      & $4.407 \pm 1.982$ & $\bm{4.337 \pm                            0.253}$ \T               \\  
\bottomrule
\end{tabular}


\end{table*}

\section{Experiments}
\label{sec:exp}
We first empirically show the accuracy and robustness of NDO by comparing the estimated and true derivatives of various dynamical systems.
Then, we mainly consider three different types of tasks to show the advantages of proposed NDO-NODE: and i) physical systems, ii) stiff ODEs, iii) real-world airplane vibration dataset.

We use these three classes of experiments to show that:  i)  NDO-NODE improves the interpolation and extrapolation accuracy of vanilla NODE and regularized NODE; 
ii) the enhanced derivative signal helps neural ODEs to better capture the stiff transitions for stiff or vibrated dynamics, which vanilla neural ODEs are hard to capture; and iii) NDO-NODE is robust to noise and the hyperparameter $\lambda$.

\subsection{Interpolation and Extrapolation Accuracy}
\label{sec:physical}
In this section, we report the interpolation and extrapolation accuracy of NDO-NODE on physical dynamics including Linear Planar Spiral Systems, Damped Harmonic Oscillator and Three-body Problem. 

\textbf{Planar Spiral Systems} is described by two dimensional linear ODEs, which is written as 
\begin{align}
    \begin{cases} \frac{\diff x}{\diff t}= ax + by \\ \frac{\diff y}{\diff t}=cx+dy ~. \end{cases} \normalsize
\end{align}
 In our experiments, we set $a= -0.1, b = 2, c = -2, d=-0.1$ and initial values $[x,y] = [2,0]$ \cite{chenNeuralOrdinaryDifferential2018}.
 
\textbf{Damped Harmonic Oscillator} is a vibrating system whose amplitude of vibration decreases over time \cite{goldstein2002classical}. It is described as
\begin{align}
 \begin{cases} \frac{\diff x}{\diff t}= v \\ \frac{\diff v}{\diff t}= -\left(\omega^{2}+\gamma^{2}\right) x-2 \gamma v ~, \end{cases}
\end{align}
where $x,v$ are the position and velocity of the oscillator, and $\omega$, $\gamma$ describe the undamped angular frequency and damping coefficient. We set $\gamma=0.1$, $\omega=1$ \cite{norcliffeSecondOrderBehaviour2020}. 

\textbf{Three-body Problem} 
is a chaotic physical system, whose motion is governed by
\begin{align}  
    \frac{\diff ^2 \mathbf{x}_i}{\diff t^2}=-\sum_{j \neq i} G m_{j} \frac{\mathbf{x}_{i}-\mathbf{x}_{j}}{|\mathbf{x}_{i}-\mathbf{x}_{j}|^{3}}, \qquad  i=1,2,3 ~, \normalsize
\end{align} 
where $\mathbf{x}_i$ denotes the position of $i$th body in 3-dimensional space, $m_i$ denotes the mass of $i$th body, and $G$ stands for gravitational constant. 

We choose vanilla NODE \cite{chenNeuralOrdinaryDifferential2018} and RNODE \cite{finlayHowTrainYour2020a} as our baselines, because RNODE directly constrains the $L_2$ norm of derivatives $f_\Theta$ to zero, which is most close to our methods. For each experiment, we select the best coefficient $\lambda$ of the regularization term from the range $ \{10^{-4}, 10^{-3}, \cdots, 1\}$ for NDO-NODE and RNODE.
We measure the accuracy by the mean squared error of predictions with respect to ground truth.

In the simulated experiments, we divide the whole time range $[0,T]$ into two segments, $[0,T_1]$ and $[T_1,T_2]$. We train our model on $[0,T_1]$ and validate on both $[0,T_1]$ and $[T_1,T_2]$ for interpolation and extrapolation, respectively. For training data, we irregularly choose $100$ time points from the training time range $[0,T_1]$, and generate the corresponding process states.
For the test data, we uniformly choose $1000$ time points from $[0,T_1]$ and $[T_1,T_2]$, respectively. We measure the interpolation ability by the mean squared error on $[0,T_1]$ (In. MSE), and 
extrapolation ability by the mean squared error on $[T_1,T_2]$  (Ex. MSE). 
Training details can be found in Appendix.

We report the interpolation and extrapolation mean squared error (MSE) for the three physical systems in Table \ref{table:threebody}. The results show that NDO-NODE achieves the lowest MSE compared with NODE and RNODE on both interpolation and extrapolation. It indicates that   auxilary signal provided by NDO can help to learn the dynamics more accurate. Due to space limitation, we put further demonstrations on the learned dynamics trajectories and results on these dynamics with perturbed noise in Appendix.

\subsection{Performance on Dynamics with Sharp Transitions}
\subsubsection{Stiff ODEs}
\begin{wrapfigure}{RH}{5.5cm}
\centering
\includegraphics[width=5.5cm]{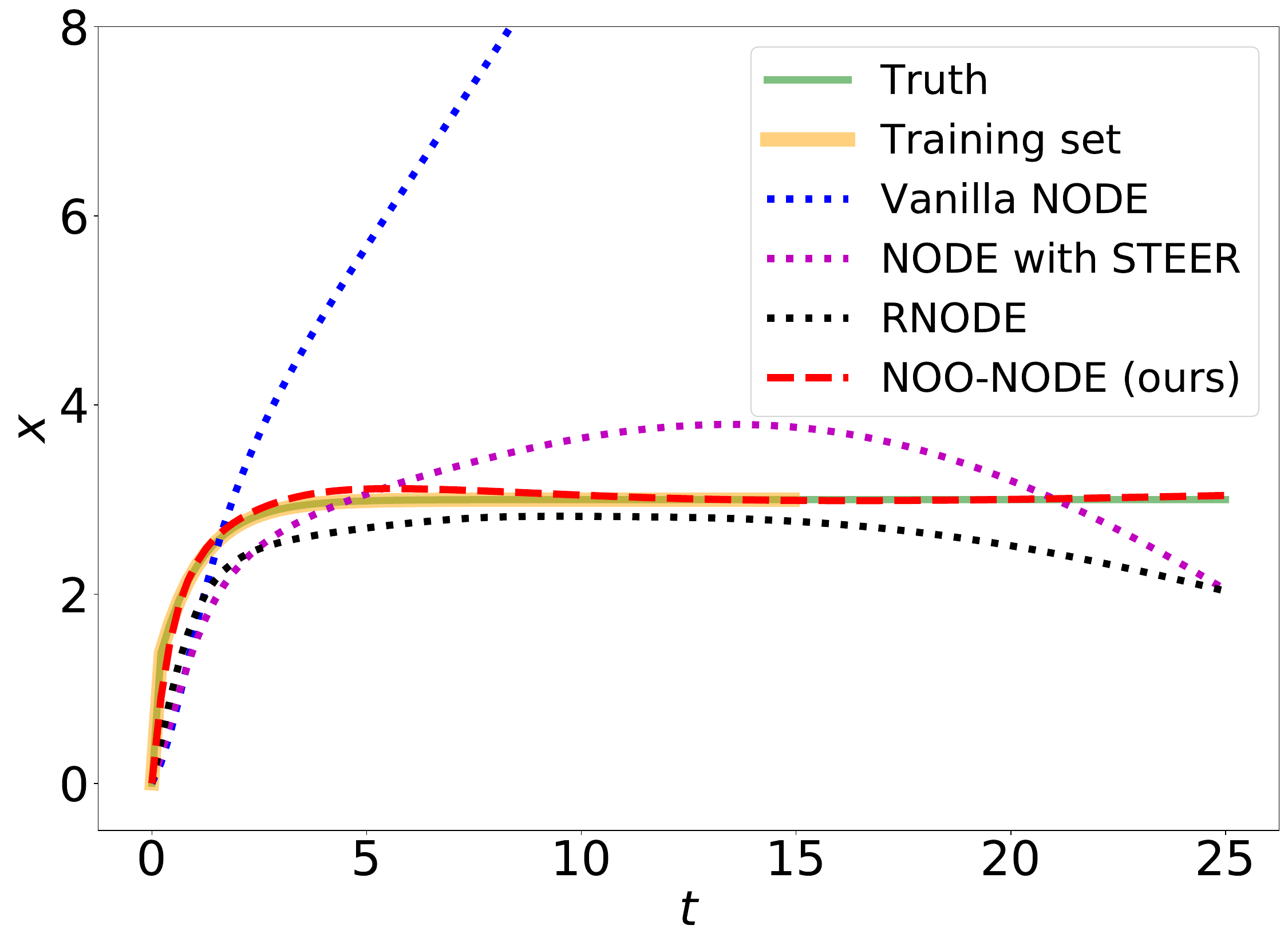}
  \caption{Results on the stiff ODE}
  \vspace{-0.3cm}
  \label{fig:stiff}
\end{wrapfigure}
We select the 
 stiff ODE in \citet{ghosh2020steer} i.e.,
\begin{align}
    \frac{dx}{dt} &= -1000x+3000-2000e^{-t} \label{ode:plane}  
\end{align}
with initial condition $x(0)=0$ to study.  We use the same setting for vanilla NODE and STEER as \citet{ghosh2020steer} do
in their released code, and add derivative signal with other parameters unchanged for NDO-NODE. 
We train these models on time range $[0,15]$ while forecast on range $[15,25]$. 
The learned trajectories are shown in Figure~\ref{fig:stiff}. 
It shows that vanilla NODE fails to learn this system and deviates far from the ground truth, while NODE with STEER and RNODE performs better but is still not accurate. NDO-NODE learns this system well and captures the evolution of this dynamics. This result indicates that the smoother estimation on the stiff point and the captured overall tendency of the derivatives (as shown in Figure 4(c)) of NDO helps the training of NODE.

\subsubsection{Airplane Vibration Dataset}

\begin{figure}[h]
  \centering
  \includegraphics[width=0.45\linewidth]{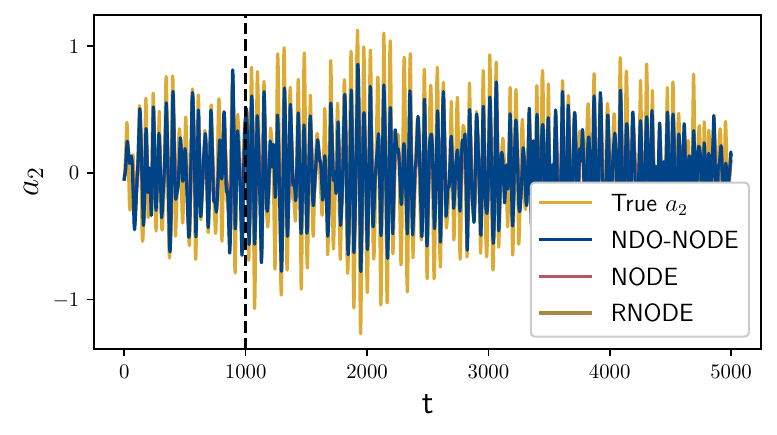}
  \includegraphics[width=0.45\linewidth]{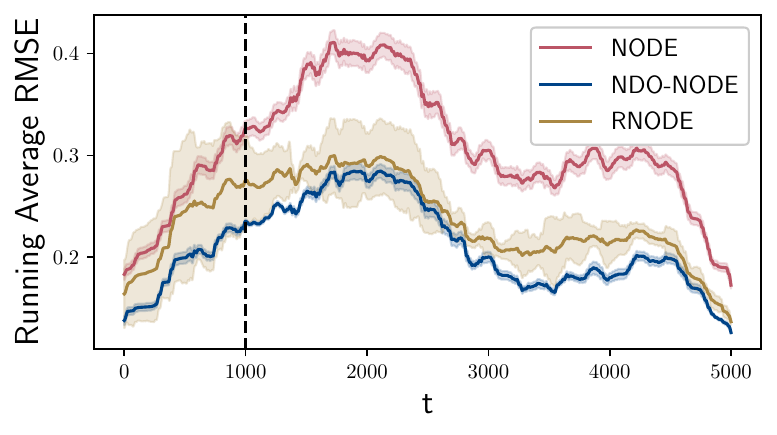}
  \caption{Results on airplane vibration dataset \textbf{(Left)}: ground truth and forecasting trajectories. \textbf{(Right)}: moving averages of root mean squared error (RMSE).}
  \label{fig:airplane}
\end{figure}

This dataset \cite{noel2017f} records the acceleration signals of an aircraft based on a ground vibration test. As shown in the top figure in Figure~\ref{fig:airplane}, its dynamic is vigorously shaking. 
\footnote{$a_2$ in Figure~\ref{fig:airplane} is measured on the right wing next to the nonlinear interface of interest.} 

Similar to \cite{norcliffeSecondOrderBehaviour2020}, we test vanilla NODE, RNODE and NDO-NODE on this dataset. The networks in above three methods are parameterized as a two fully connected layers. We train the models on $[0,1000]$ and forecast on $[1000,5000]$ time units. The results are reported in the bottom one in Figure~\ref{fig:airplane}, which shows NDO-NODE has the lowest RMSE over the others on this dataset. The predicted dynamics are also reported in the above one in  Figure~\ref{fig:airplane}. We can observe that the output of NDO-NODE is more closed with the ground truth $a_2$.\footnote{The outputs for NODE RNODE are almost covered by NDO-NODE because their outputs can not capture the sharply vibration. Further demonstrations about this result are put in Appendix.}

\subsection{Ablation Study}
\begin{figure*}[t]
 	\centering
 	\subfigure[\stackanchor{ \stackanchor{Different noise scales $\sigma$,}{same library parameter $P=50$}}{\textbf{(left)}: In. MSE Ratio ~ \textbf{(right)}: Ex. MSE Ratio} ]{\includegraphics[width=0.24\textwidth]{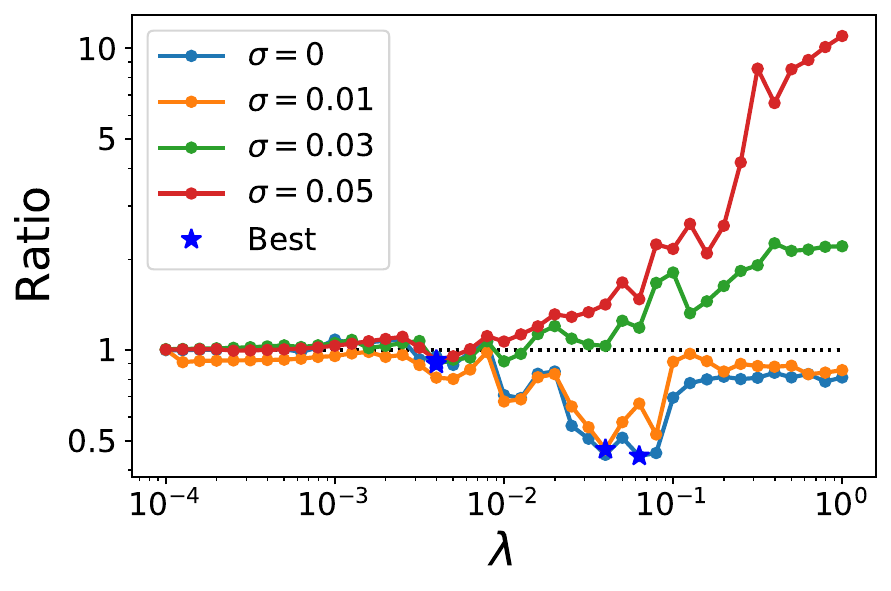} \includegraphics[width=0.24\textwidth]{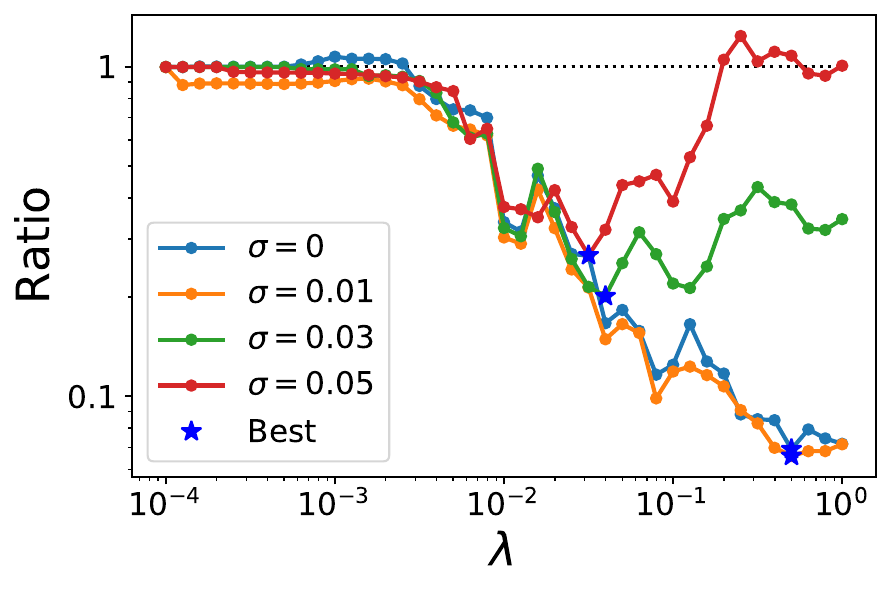}\label{fig:ratio_noise}} 
	\subfigure[\stackanchor{\stackanchor{Different Library parameter $P$,}{same noise scale $\sigma=0$}}{\textbf{(left)}: In. MSE Ratio ~ \textbf{(right)}: Ex. MSE Ratio}]{\includegraphics[width=0.24\textwidth]{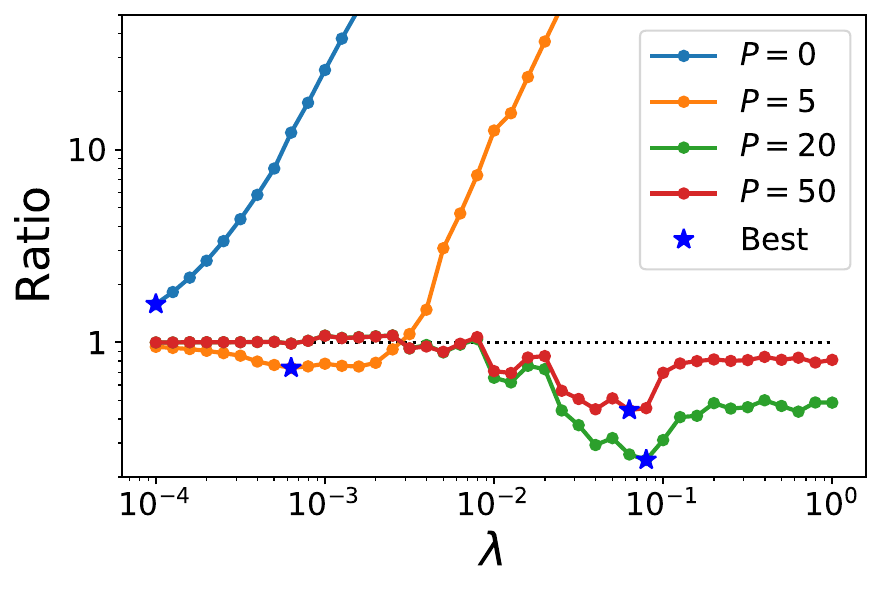} \includegraphics[width=0.24\textwidth]{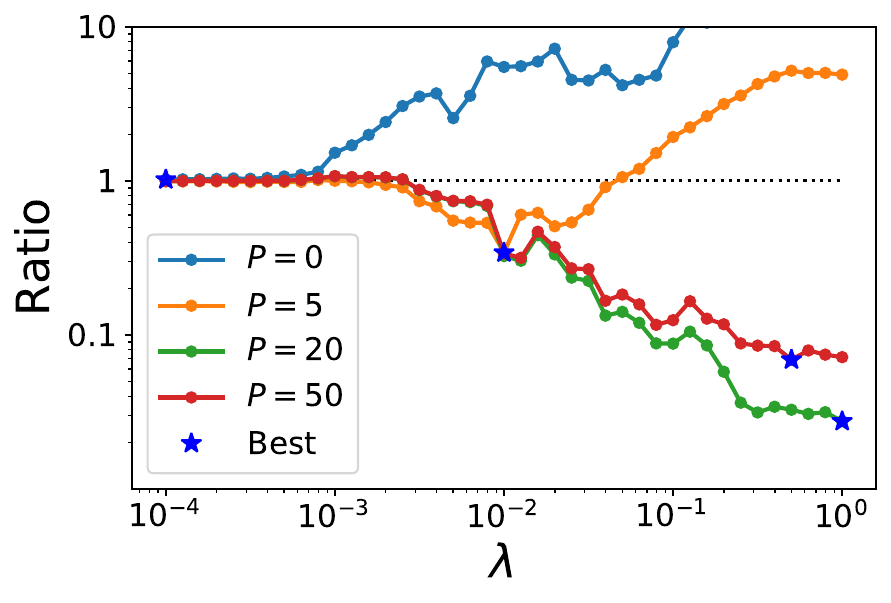}\label{fig:ratio_library}}
 	\caption{The interpolation and extrapolation ratios of NDO-NODE to Vanilla NODE ($\frac{\text{MSE of NDO-NODE}}{\text{MSE of Vanilla NODE}}$) with different signal strength $\lambda$, noise scales $\sigma$, and library complexity $P$ in planar spiral system task (see Section~\ref{sec:physical}), computed by the mean MSE over 3 runs.}
	\label{fig:ratio}
\end{figure*}

In this section, we conduct ablation studies to provide a better understanding of how signal strength $\lambda$, library complexity, and noise in data sampling affect the performance of NDO-NODE. To test the robustness under different noise scales, as \citet{norcliffeSecondOrderBehaviour2020} do, we add noises that are independently drawn from $\mathcal{N}(0,\sigma^2)$ to each state in training data.

Taking the planar spiral system task in Section~\ref{sec:physical} as an example, we grid search $\lambda$ in $[10^{-4},1]$ for different libraries and noises. 

\subsubsection{Noise Robustness}
We test NDO-NODE by adding noises $\sigma=0,0.01,0.03,0.05$ to the training data and keeping the library of NDO unchanged with $(P,Q,C)=(50,3,10)$. The results are shown in Figure~\ref{fig:ratio_noise}, from which we observe that NDO-NODE is robust to noise and can consistently improve the interpolation and extrapolation accuracy under different noise scales. Among all the best $\lambda$s of interpolation and extrapolation, the lower the noise scale, the more significant improvement NDO-NODE achieves. Due to NDO outputs more accurate derivative estimations under lower noise scales, these observations are consistent with the intuition that, the enhanced derivative signal provides stronger guidance when the estimations are more accurate.

\subsubsection{Signal strength \texorpdfstring{$\lambda$}{Lg}}
The results in Figure~\ref{fig:ratio_noise} suggest that it is better to choose a relatively large $\lambda$ under the low noise scales for greater improvements, and vice versa. When $\lambda$ is large, NDO tends to provide a stronger supervised signal to NODE, which helps extrapolation because NDO is expected to capture the overall curvature information. However, the derivative estimations form NDO can not be exactly precise, and thus when we use large $\lambda$, the interpolation MSE may get larger even under low noise scale.    

\subsubsection{Library Complexity}
We test NDO-NODE with different libraries, i.e., $P=0,5,20,50$ and $Q=3,C=10$, under noise $\sigma = 0$, and the results are shown in Figure~\ref{fig:ratio_library}. The results indicate NDO is helpful for interpolation and extrapolation even on a small library $P=5$, and a larger library will lead to better results. When the library is extremely small such as $P=0$, it can be hard on NDO to to generalize to other functions. These results also match our observations in Section~\ref{sec:ndo}.
 
\section{Conclusion}\label{sec:con}
We propose an algorithm called NDO-NODE, which leverages the estimated derivatives from trajectory samples, to enhance the supervised signal of the NODE training process. The estimated derivatives are obtained by the neural differential operator, which is pre-trained on a class of basis functions. With the supervision of estimated derivatives, NDO-NODE can improve the forecasting accuracy on various dynamics. 
We believe that this work starts a new direction on dynamics pre-training and there are a lot of interesting research topics for future study. First, for the selection of the pre-training data, we can explore the more structured libraries of functions such as ODEs and PDEs to pre-train the operator. Second, for the training mechanism, we can also let the pre-trained model as the initialization for downstream tasks, which we will investigate in the future. Third, training an NDO for functions with multi-dimensional input is also an important direction. It is challenging due to the discretization is impacted by the curse of dimensionality. We may consider some Monte Carlo based methods to overcome this difficulty.



\bibliographystyle{abbrvnat}
\bibliography{aaai22}

\clearpage
\appendix



\section{Experimental Details}
All experiments are performed with Python 3.6 and PyTorch 1.8.1. We use differentiable ODE solver \footnote{See their Github repo at \url{https://github.com/rtqichen/torchdiffeq}.} implemented by \citet{chenNeuralOrdinaryDifferential2018}, and we choose the adaptive step size solver \texttt{Dopri5} by default. Neural differential operator is trained on a single NVIDIA Tesla P100 GPU, and other experiments are on a single CPU. RNODE in our baseline regularizes the derivatives directly to zero, i.e., the loss function is $\mathcal{\widetilde{L}} = \mathcal{L}(\mathcal{X}',\mathcal{X}) + \lambda \cdot\| f_\Theta (\mathcal{X},\mathcal{T}) \|_2^2$.

\subsection{Neural Differential Operator} \label{A1}
We implement all NDOs in following experiments by a 2-hidden-layer bidirectional LSTM followed by an output layer. Each hidden layer of LSTM has 128 units, and the output layer is defined as a 3-layer fully connected network 128-64-32-1 with ReLU activation. 
We randomly draw $10000$ function samples from $\mathcal{Z}_{lib}$ and discretize them by $100$ uniform random times in interval $[0,1]$ as our training data. For the training process, we use Adam optimizer with an initial learning rate of $0.003$ and decayed by \texttt{CosineAnnealingLR} scheduler. The minibatch size is set to 64 and we train for $641$ epochs ($100000$ iterations).

For the first-order NDO, we set the input sequence as $(\mathcal{X},\mathcal{T},\Delta\mathcal{T}) = \left\{(x_i,t_i,\Delta t_i)\right\}_{i=0}^N$, where $\Delta t_i = t_i-t_{i-1}$, and the corresponding labels as $\dot{\mathcal{X}}$. For the second-order NDO,
we set the input sequence as $(\dot{\mathcal{X}},\mathcal{X},\mathcal{T},\Delta\mathcal{T}) = \left\{(\dot{x_i},x_i,t_i,\Delta t_i)\right\}_{i=0}^N$, where $\Delta t_i = t_i-t_{i-1}$, and the corresponding labels as $\ddot{\mathcal{X}}$.

For the airplane vibration dataset, due to the number of training time points is $1000$, we cut them into $10$ segments with $100$ time points in each segment, to adapt to the above NDOs trained on $100$ time points.

In the three-body problem, a position vector $\mathbf{r}(t)$ in 3-dimensional space can be written in the parametric form $\mathbf{r}(t) = \left(r_1(t),r_2(t),r_3(t) \right)$. Thus, we define the input of the first-order NDO as $(\mathcal{X}_1,\mathcal{X}_2,\mathcal{X}_3,\mathcal{T},\Delta\mathcal{T})  = \left\{(x_{1i},x_{2i},x_{3i},t_i,\Delta t_i)\right\}_{i=0}^N$, where $\mathcal{X}_1,\mathcal{X}_2,\mathcal{X}_3$ can be generated from three independent function samples drawn from the library $\mathcal{Z}_{lib}$. The corresponding label is  $(\dot{\mathcal{X}}_1,\dot{\mathcal{X}}_2,\dot{\mathcal{X}}_3)$. The second-order NDO can be similarly defined and trained for this task.

\subsection{Experiments on Physical Systems}
\subsubsection{Planar Spiral Systems}
\begin{table*}[!htbp]
\centering
\caption{Mean squared error (MSE) (mean $\pm$ std, $\times 10^{-2}$, over 3 runs) in planar spiral systems experiments. In. MSE (interpolation MSE) is computed on time range $[0,5]$ seconds while Ex. MSE (extrapolation MSE) is computed on time range $[5,10]$ seconds. }
\label{table:spiral}
\begin{tabular}{ccccc}
\toprule
MSE & Noise & NODE & RNODE & NDO-NODE (ours) \\
\midrule 
\multirow{4}{*}{In. MSE} & 0                      & $0.026 \pm 0.014$     & $0.040 \pm 0.019$     & $\bm{0.012 \pm 0.004}$                \\
                          & 0.01                   & $0.036 \pm 0.017$     & $0.049 \pm 0.028$     & $\bm{0.018 \pm 0.014}$                \\
                          & 0.03                   & $0.054 \pm 0.028$     & $0.069 \pm 0.040$     & $\bm{0.049 \pm 0.043}$                \\
                          & 0.05                   & $0.089 \pm 0.041$     & $0.101 \pm 0.053$     & $\bm{0.085 \pm 0.039}$                \\ \midrule 
 \multirow{4}{*}{Ex.MSE}  & 0                      & $4.524 \pm 2.890$     & $4.023 \pm 2.749$     & $\bm{0.525 \pm 0.304}$                \\
                         & 0.01                   & $5.416 \pm 3.671$     & $4.788 \pm 3.513$     & $\bm{0.530 \pm 0.371}$                \\
                          & 0.03                   & $5.717 \pm 3.652$     & $5.174 \pm 3.479$     & $\bm{1.853 \pm 1.845}$                \\
                          & 0.05                   & $6.332 \pm 3.902$     & $5.729 \pm 3.671$     & $\bm{5.347 \pm 4.193}$                \\ \bottomrule 
 \end{tabular}
 \end{table*}
\begin{figure*}[!htbp]
 	\centering
 	\subfigure[$\sigma=0$]{\includegraphics[width=0.24\textwidth]{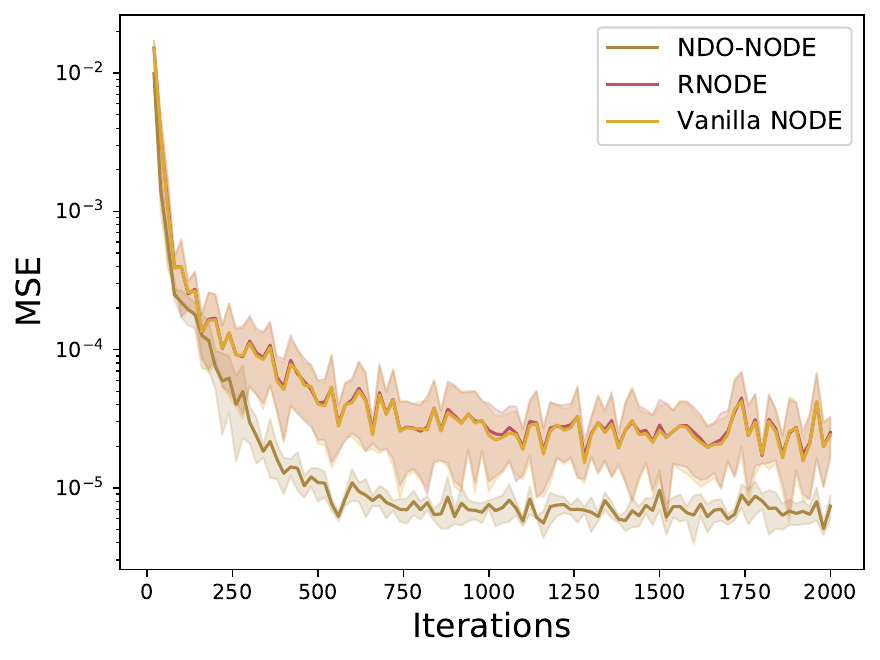}} 
	\subfigure[$\sigma=0.01$]{\includegraphics[width=0.24\textwidth]{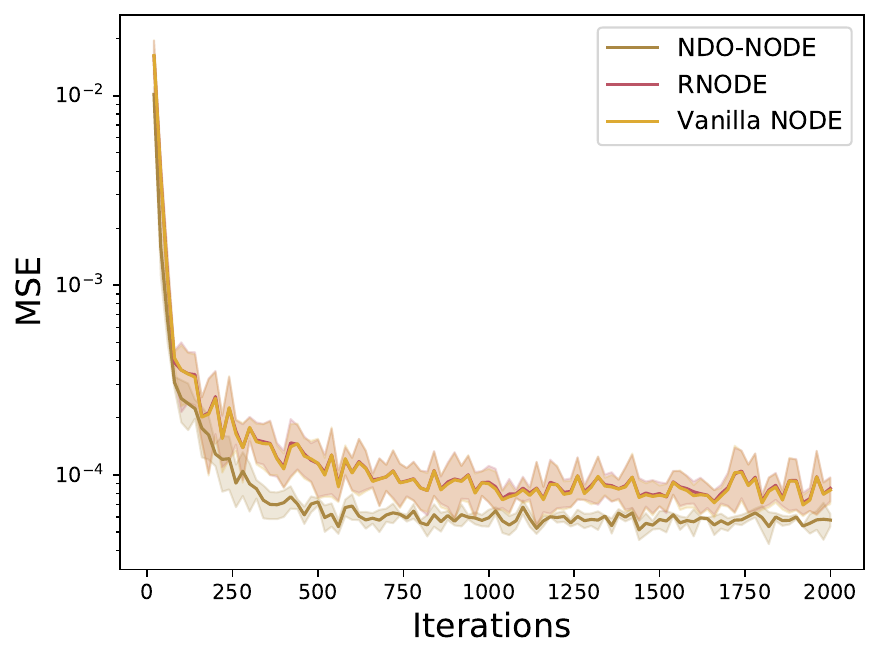}}
	\subfigure[$\sigma=0.03$]{\includegraphics[width=0.24\textwidth]{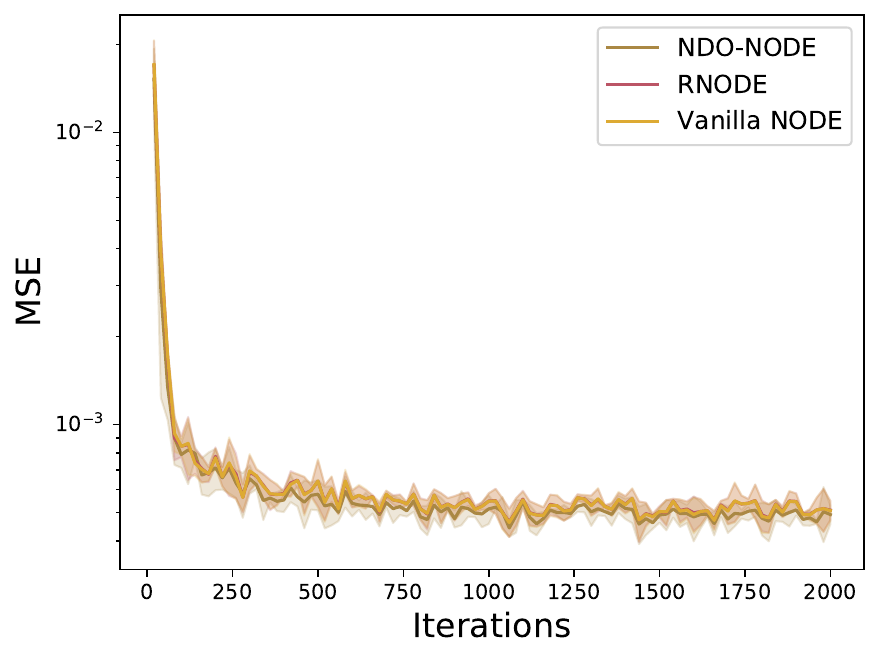}}
	\subfigure[$\sigma=0.05$]{\includegraphics[width=0.24\textwidth]{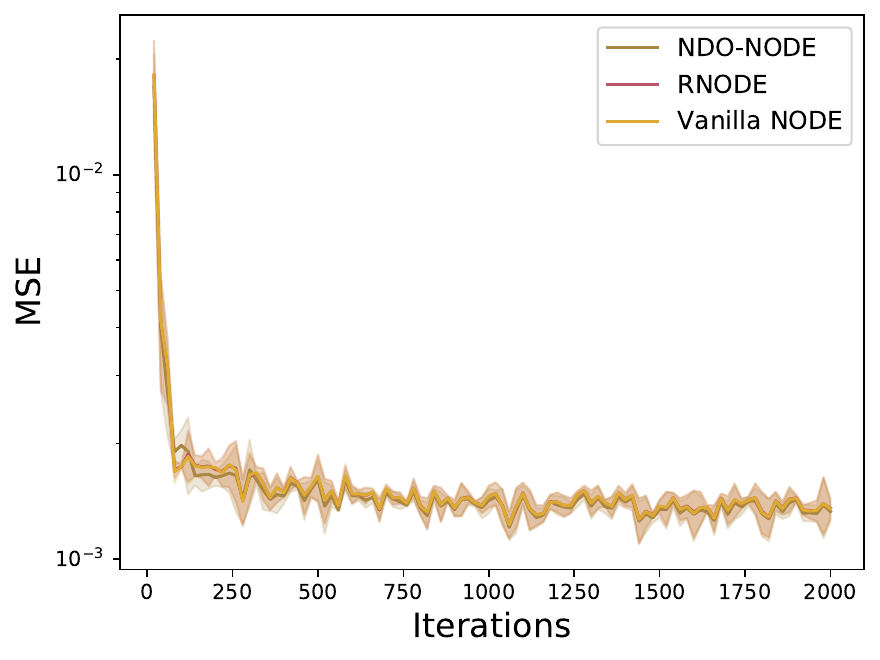}}
 	\caption{Training MSE loss of planar spiral systems under different noise scales.} 
 	\label{fig:loss_spiral}
\end{figure*}
Linear ordinary differential systems are one of the fundamental equations and are widely used in physics \cite{4452}. We take two dimensional linear ODEs as an example, which can be written as
\begin{align}
    \begin{cases} \frac{\diff x}{\diff t}= ax + by \\ \frac{\diff y}{\diff t}=cx+dy ~. \end{cases} \normalsize
\end{align}
\citet{chenNeuralOrdinaryDifferential2018} studies the system with parameters $a= -0.1, b = 2, c = -2, d=-0.1$ and initial values $[x,y] = [2,0]$, whose trajectory looks like a planar spiral coil. Following their setting, we parameterize a 2-dimensional neural ODEs with a one hidden layer network with 20 hidden units and ELU activation \cite{clevert2015fast}. We set the training time range to $[0,5]$ seconds and test extrapolation on $[5,10]$ seconds, and Table~\ref{table:spiral} shows the results under different noise scales. NDO-NODE performs consistently better than baselines on both interpolation and extrapolation tasks, and it shows prominent advantages under low noise scale since NDO provides more accurate estimation of derivatives. Figure~\ref{fig:spiral} visualizes how the learned dynamics extrapolate and the help on extrapolation from derivative signal. 

\label{sec:spiral}
\begin{figure}[h]
	\centering
 	\subfigure[NDO-NODE]{\includegraphics[width=0.2\textwidth]{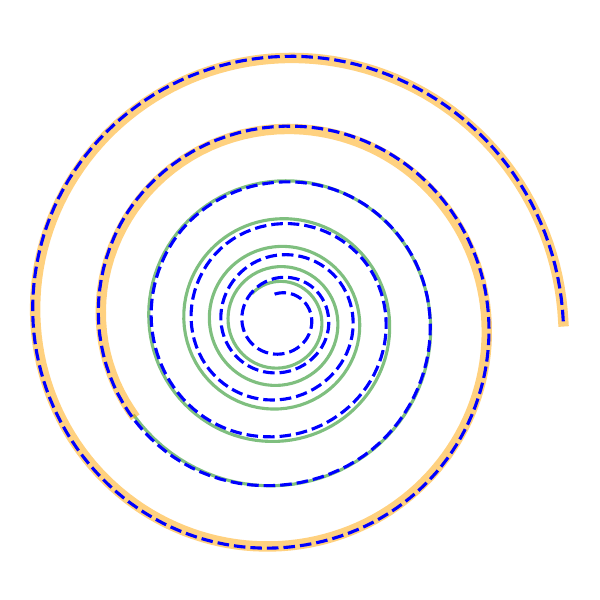}}
	\subfigure[Vanilla NODE]{\includegraphics[width=0.2\textwidth]{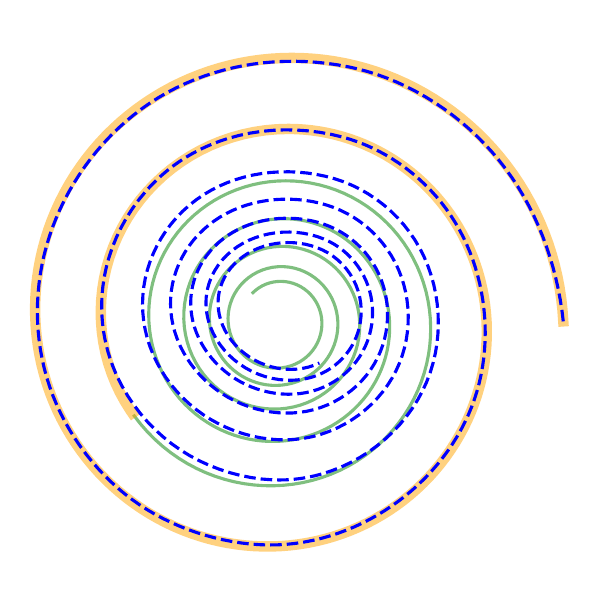}}
 	\caption{Ground truth (\textcolor{xgreen2}{green soild}), and prediction (\textcolor{xblue2}{dashed blue}) of spiral systems under noise $\sigma = 0$. Training on $[0,5]$ seconds (\textcolor{xorange2}{bold orange}) and forecasting on $[5,20]$ seconds.}
 	\label{fig:spiral}
\end{figure}

We train all models for 2000 iterations by Adam optimizer with an initial learning rate of 0.1 and decayed by a factor of 0.995 at each iteration.  The state of the ODEs is defined as $[x,y]$, and we apply first-order NDO separately on the trajectories $(x_i)_{i=1}^N$ and $(y_i)_{i=1}^N$ to get corresponding estimated derivatives $(\dot{x}_i)_{i=1}^N$ and $(\dot{y}_i)_{i=1}^N$. For NDO-NODE and RNODE, we grid search the strength $\lambda$ in range $[10^{-4},1]$ for the best performance. We choose $\lambda = 0.08, 0.08,0.01, 0.005$ for NDO-NODE under noise scales $\sigma = 0,0.01,0.03,0.05$, respectively, and $\lambda = 0.0001, 0.0001, 0.0001, 0.0001$ for RNODE. Figure \ref{fig:loss_spiral} shows the training MSE loss of these models under different noise scales.

\subsubsection{Damped Harmonic Oscillator}
\label{sec:damped}
The damped harmonic oscillator is a vibrating system whose amplitude of vibration decreases over time. It is a typical model in physics that has been widely studied \cite{goldstein2002classical}.  From mechanics, this system can be governed by following ordinary differential systems
\begin{align}
 \begin{cases} \frac{\diff x}{\diff t}= v \\ \frac{\diff v}{\diff t}= -\left(\omega^{2}+\gamma^{2}\right) x-2 \gamma v ~, \end{cases}
\end{align}
where $x,v$ are the position and velocity of the oscillator, and $\omega$, $\gamma$ describe the undamped angular frequency and damping coefficient. We set $\gamma=0.1$, $\omega=1$ and generate 30 random position trajectories with different initial positions and velocities, under the similar setting of \citet{norcliffeSecondOrderBehaviour2020}. The state $[x,v]$ is modeled by a one hidden layer neural ODEs with 20 hidden units. As we only have the position trajectory, we use both $1$st-order and $2$nd-order NDO to extract the underlying velocity $v$ and the acceleration $a=\frac{\diff v}{\diff t}$. We train the models on $[0,10]$ seconds, and forecast the positions on $[10,20]$ seconds.  The results under different noise scales are listed in Table~\ref{table:oscillator}. NDO-NODE provides a big improvement to both interpolation and extrapolation under low scales. When noise gets higher, NDO output inaccurate derivative estimation, which provides less information comparing to low noise scales.

\begin{figure*}[!htbp]
 	\centering
 	\subfigure[$\sigma=0$]{\includegraphics[width=0.24\textwidth]{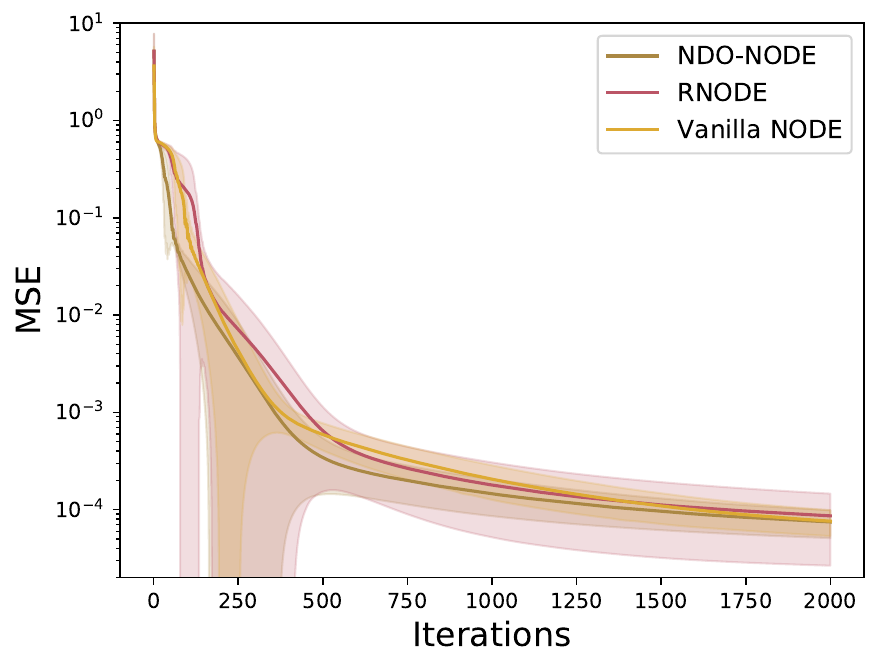}} 
	\subfigure[$\sigma=0.1$]{\includegraphics[width=0.24\textwidth]{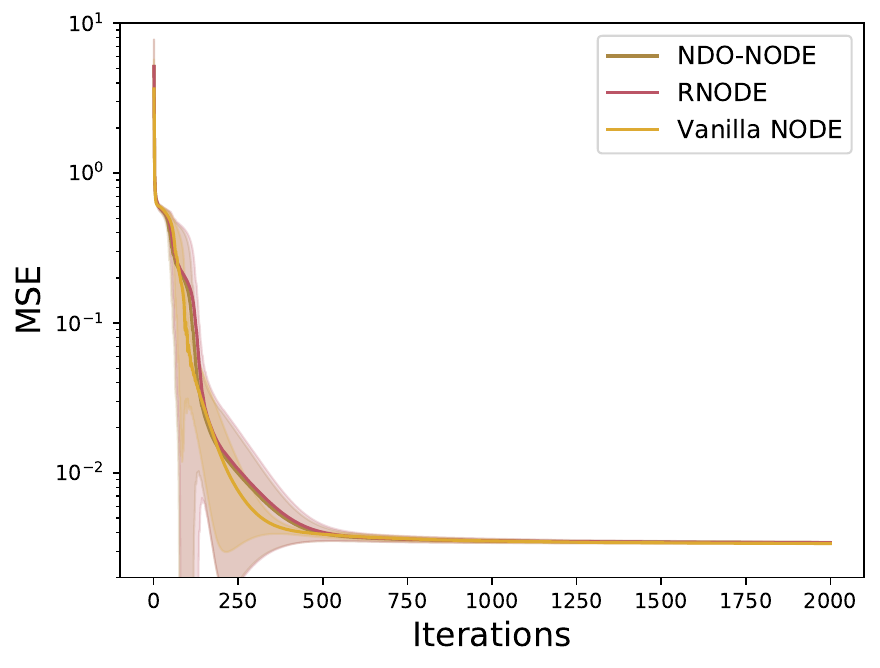}}
	\subfigure[$\sigma=0.3$]{\includegraphics[width=0.24\textwidth]{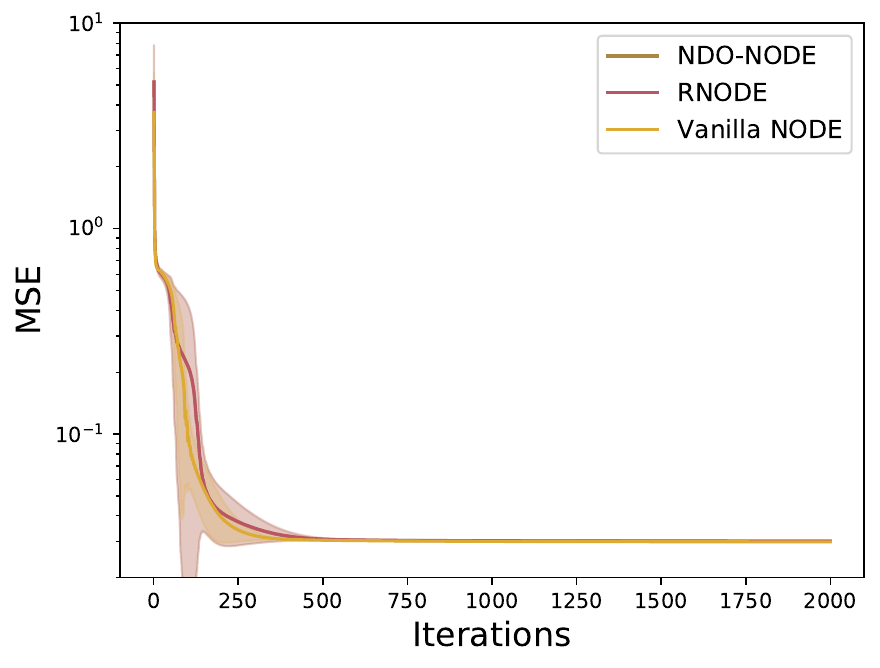}}
	\subfigure[$\sigma=0.5$]{\includegraphics[width=0.24\textwidth]{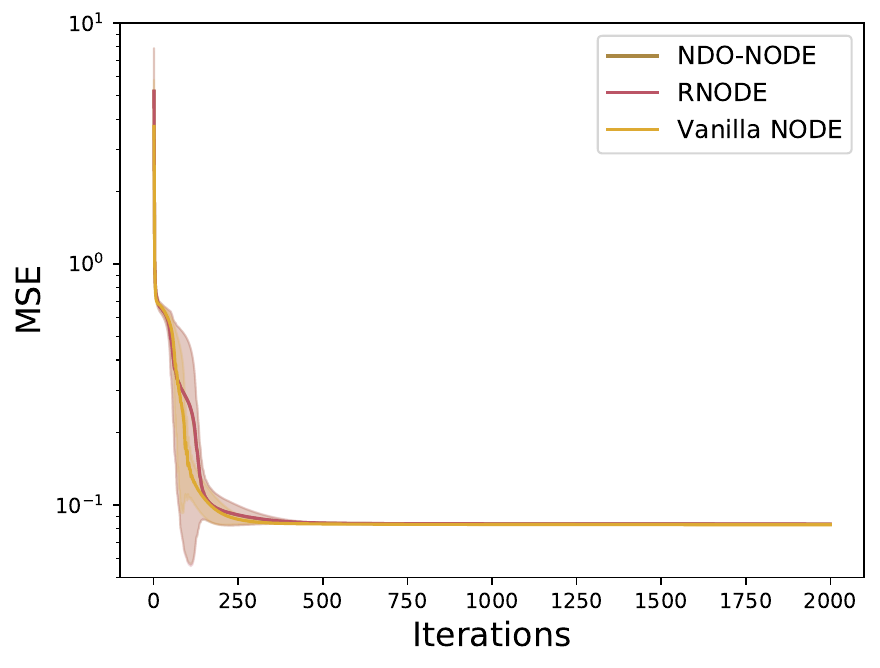}}
 	\caption{Training MSE loss of damped harmonic oscillator under different noise scales.} 
 	\label{fig:loss_osci}
\end{figure*}

Similar to the setting in \citet{norcliffeSecondOrderBehaviour2020}, we train all models for 2000 iterations by Adam optimizer with an initial learning rate of 0.01 and decayed by a factor of 0.999 at each iteration. Let $x$, $v = \frac{\diff x}{\diff t}$ and $a = \frac{\diff v}{\diff t}$ denotes the position, velocity and acceleration of the oscillator, respectively. We model the state of the ODEs as $[x,v]$. To get the corresponding estimated derivative $[v,a]$, we use first-order NDO to extract velocity estimations $(v_i)_{i=1}^N$ from position observations $(x_i)_{i=1}^N$, and then use second-order NDO to extract acceleration estimations $(a_i)_{i=1}^N$ from both observations $(x_i)_{i=1}^N$ and estimated velocities $(v_i)_{i=1}^N$.  For $\sigma = 0,0.1,0.3,0.5$, we choose $\lambda = 0.8, 0.02, 0.001, 0.001$ for NDO-NODE, and $\lambda = 0.001, 0.001, 0.01, 0.01$ for RNODE, after grid search $\lambda$ in range $[10^{-4},1]$. Figure \ref{fig:loss_osci} shows the training MSE loss of these models under different noise scales.

\begin{table*}[h]
\centering
\caption{Mean squared error (MSE) (mean $\pm$ std, $\times 10^{-4}$, over 3 runs) in damped harmonic oscillator experiments. In. MSE (interpolation MSE) is computed on time range $[0,10]$ seconds while Ex. MSE (extrapolation MSE) is computed on time range $[10,20]$ seconds. }
\label{table:oscillator}
\begin{tabular}{ccccc}
\toprule
MSE & Noise & NODE & RNODE & NDO-NODE(ours) \\
\midrule 
\multirow{4}{*}{In. MSE} & 0                      & $5.217 \pm 1.948$       & $2.058 \pm 0.872$       &$\bm{ 1.005 \pm 0.305}$                  \\
                          & 0.1                    & $5.559 \pm 2.014$       & $1.835 \pm 0.355$       & $\bm{ 1.775 \pm 0.465}$                  \\
                          & 0.3                    & $9.494 \pm 4.448$       & $3.824 \pm 1.480$       & $\bm{ 3.389 \pm 0.615}$                  \\
                          & 0.5                    & $16.67 \pm 7.856$       & $\bm{7.464 \pm 1.919}$       & $7.530 \pm 1.732$                  \\ \midrule
\multirow{4}{*}{Ex.MSE}  & 0                      & $7.918 \pm 5.762$       & $4.239 \pm 2.058$       & $\bm{ 0.860 \pm 0.336}$                  \\
                          & 0.1                    & $7.900 \pm 5.628$       & $3.099 \pm 0.961$       & $\bm{ 2.470 \pm 1.291}$                  \\
                          & 0.3                    & $12.32 \pm 7.901$       & $3.353 \pm 1.589$       & $\bm{ 2.433 \pm 1.300}$                  \\
                          & 0.5                    & $20.52 \pm 11.95$       & $4.661 \pm 2.332$       & $\bm{ 4.637 \pm 2.233}$                  \\ \bottomrule
\end{tabular}
\end{table*}

\subsubsection{Three-body Problem}
\label{sec:3body}
The three-body problem is one of the most famous and important problems in physics and celestial mechanics, which was first proposed to model the motion of three celestial bodies \cite{valtonen2006three}. By Newton's laws of motion and Newton's law of universal gravitation, this dynamical system is governed by
\begin{align}  
    \frac{\diff ^2 \mathbf{r}_i}{\diff t^2}=-\sum_{j \neq i} G m_{j} \frac{\mathbf{r}_{i}-\mathbf{r}_{j}}{|\mathbf{r}_{i}-\mathbf{r}_{j}|^{3}}, \qquad  i=1,2,3 ~, \normalsize
\end{align} 
where $\mathbf{r}_i$ denotes the position of $i$th body in 3-dimensional space, $m_i$ denotes the mass of $i$th body, and $G$ stands for gravitational constant. Because this system is chaotic for most initial conditions, we integrate partial physical prior knowledge into the neural ODEs as \citet{zhuangAdaptiveCheckpointAdjoint2020} do. Specifically, we augment the input data as an $45$-dimentional vector as 
\begin{align}
    \text{Input} = \{\mathbf{r}_i,\mathbf{r}_i -\mathbf{r}_j, \frac{\mathbf{r}_{i}-\mathbf{r}_{j}}{|\mathbf{r}_{i}-\mathbf{r}_{j}|^{1}},\frac{\mathbf{r}_{i}-\mathbf{r}_{j}}{|\mathbf{r}_{i}-\mathbf{r}_{j}|^{2}},\frac{\mathbf{r}_{i}-\mathbf{r}_{j}}{|\mathbf{r}_{i}-\mathbf{r}_{j}|^{3}} \}, j \neq i,
\end{align}
and the underlying derivative of neural ODEs are modeled by a one hidden layer network with 100 hidden units. To get the estimated derivations, we apply both $1$st-order and $2$nd-order NDO on the position of each body trajectory.
We train the models on $[0,1]$ year, and predict the position on $[1,2]$ years. Table. \ref{table:threebody} shows the results in terms of MSE, 
From Table.\ref{table:threebody}, NDO-NODE shows good improvement to vanilla NODE. However, because this is a chaotic high dimensional system and we need to extract up to $2$nd-order derivatives from trajectory, our library may not cover the full form of the trajectory, thus the estimated derivatives can only provide limited help for the learning. Moreover, high dimension causes the derivative space very complex, thus RNODE also performs better than vanilla NODE and has a similar performance as NDO-NODE. 

\begin{figure*}[!htbp]
 	\centering
 	\subfigure[$\sigma=0$]{\includegraphics[width=0.24\textwidth]{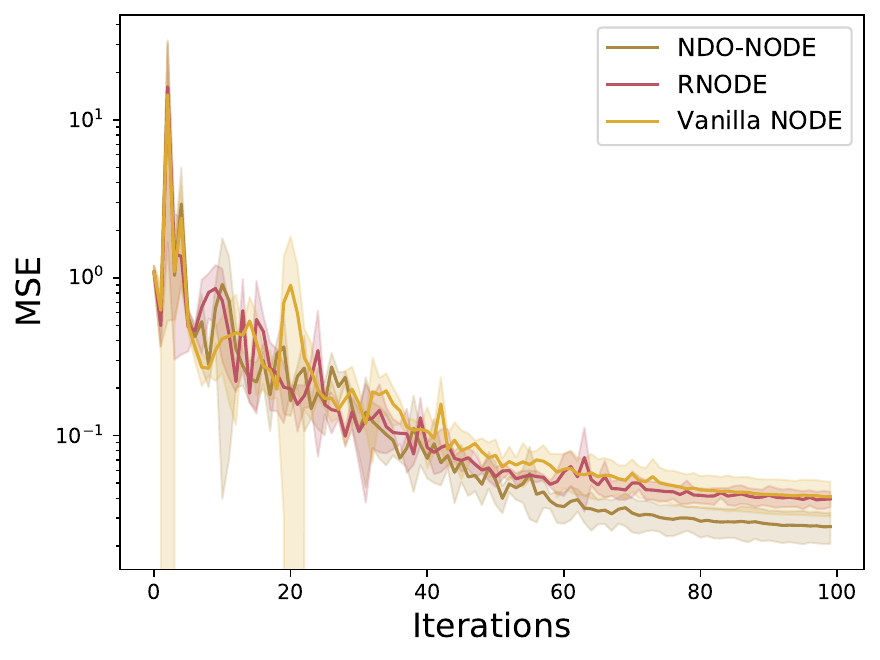}} 
	\subfigure[$\sigma=0.001$]{\includegraphics[width=0.24\textwidth]{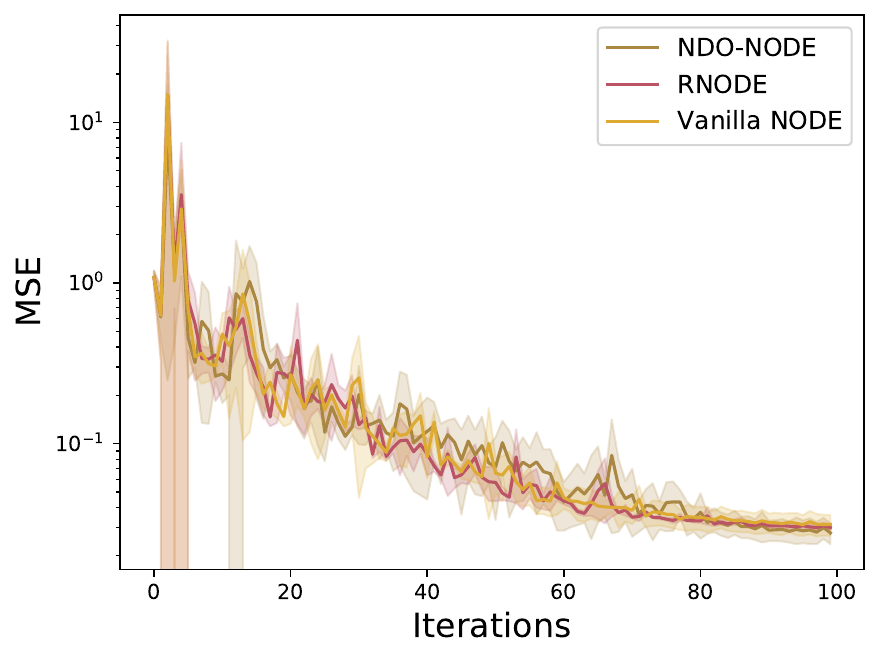}}
	\subfigure[$\sigma=0.003$]{\includegraphics[width=0.24\textwidth]{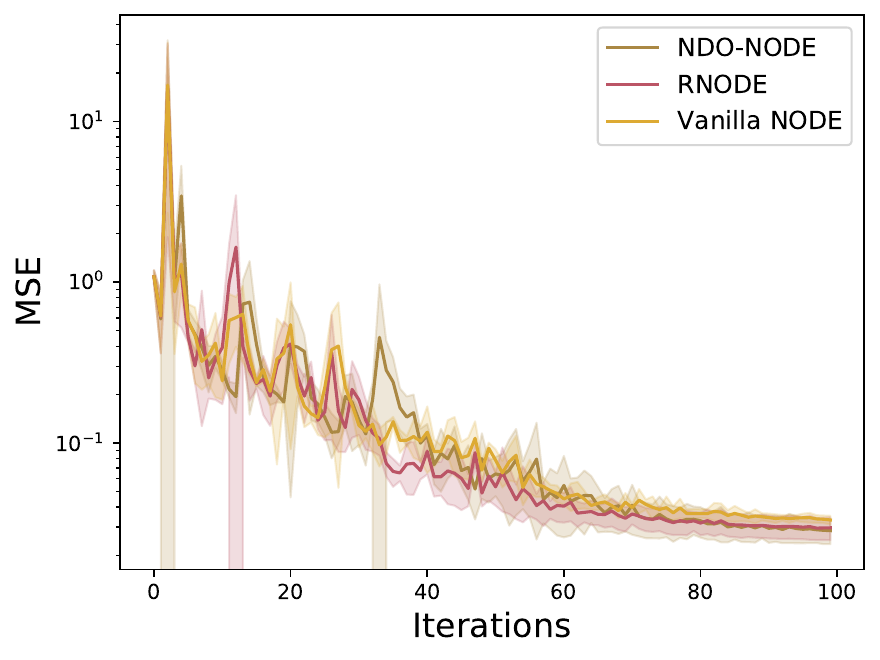}}
	\subfigure[$\sigma=0.005$]{\includegraphics[width=0.24\textwidth]{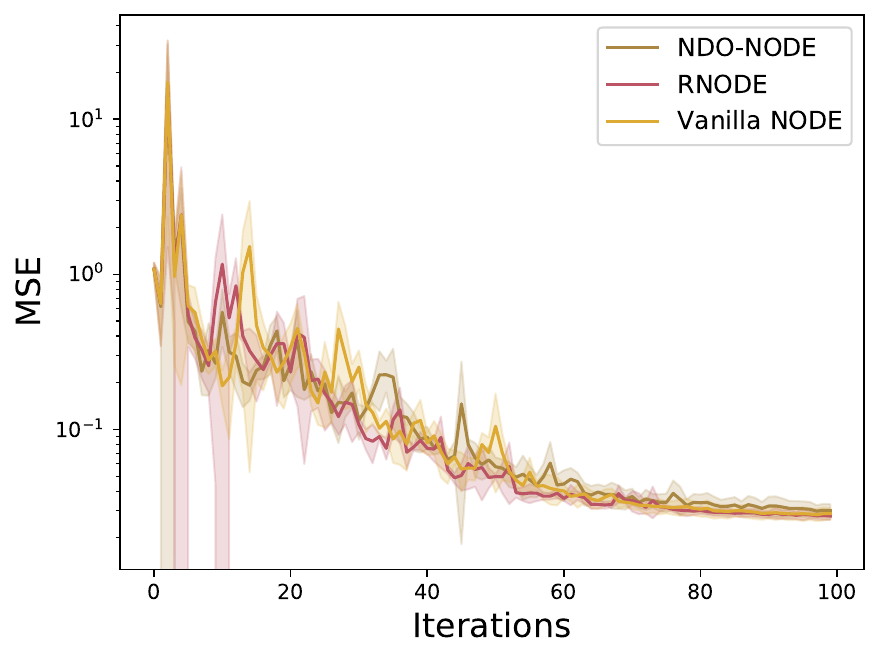}}
 	\caption{Training MSE loss of three-body problem under different noise scales.} 
 	\label{fig:loss_threebody}
\end{figure*}

Similar to the setting in \citet{zhuangAdaptiveCheckpointAdjoint2020}, we train all models for 100 iterations by Adam optimizer with an initial learning rate of 0.1 and decayed by a factor of 0.995 at each iteration. Let $\mathbf{r}_i$, $\mathbf{v}_i = \frac{\diff \mathbf{r}_i}{\diff t}$, $\mathbf{a}_i = \frac{\diff \mathbf{v}_i}{\diff t}$, $i=1,2,3$, denote the position, velocity and acceleration of $i$th body in 3-dimentional space, respectively. We model the state of ODEs as a 18-dimentional vector $[\mathbf{r}_1,\mathbf{r}_2,\mathbf{r}_3,\mathbf{v}_3,\mathbf{v}_3,\mathbf{v}_3]$. We use NDOs separately on  each body to get the estimated derivatives, as mentioned in Section \ref{A1}. For the $i$th body, we use the first-order NDO on position observations $(\mathbf{r}_{ij})_{j=0}^N$ to get velocity estimations $(\mathbf{v}_{ij})_{j=0}^N$, and then use second-order NDO to esimate the acceleration $(\mathbf{a}_{ij})_{j=0}^N$ by positions and velocities.  For $\sigma = 0,0.001,0.003,0.005$, we choose $\lambda = 0.0008, 0.004, 0.0008, 0.0015$ for NDO-NODE, and $\lambda = 0.1, 0.007, 0.002, 0.004$ for RNODE, after grid search $\lambda$ in range $[10^{-4},1]$. Figure \ref{fig:loss_threebody} shows the training MSE loss of these models under different noise scales.

\begin{table*}[h]
\centering
\caption{Mean squared error (mean $\pm$ std, $\times 10^{-1}$, over 3 runs) in the three-body problem experiments. In. MSE (interpolation MSE) is computed on time range $[0,1]$ year while Ex. MSE (extrapolation MSE) is computed on time range $[1,2]$ year. }
\label{table:threebody}
\begin{tabular}{ccccc}
\toprule
MSE & Noise & NODE & RNODE & NDO-NODE (ours) \\
\midrule 
\multirow{4}{*}{In. MSE} & 0                      & $0.414 \pm 0.110$       & $0.393 \pm 0.040$ & $\bm{0.274 \pm                           0.063}$                \\
                         & 0.001                    & $0.326 \pm 0.059$     & $0.312 \pm 0.033$ & $\bm{0.267 \pm 0.029}$                \\
                         & 0.003                    & $0.332 \pm 0.007$     & $\bm{0.299 \pm 0.046}$ & $0.334 \pm 0.038$                 \\
                         & 0.005                    & $0.287 \pm 0.038$     & $0.284 \pm 0.025$ & $\bm{0.271 \pm 0.039}$                 \\ \midrule
\multirow{4}{*}{Ex. MSE}  & 0                      & $6.908 \pm 2.077$      & $4.407 \pm 1.982$ & $\bm{4.337 \pm                            0.253}$                 \\
                         & 0.001                    & $9.944 \pm 4.467$     & $5.344 \pm 1.088$ & $\bm{4.673 \pm 1.336}$                 \\
                         & 0.003                    & $12.22 \pm 7.502$     & $5.130 \pm 1.029$ & $\bm{4.886 \pm 0.577}$                 \\
                         & 0.005                    & $9.854 \pm 2.602$     & $\bm{4.243 \pm 0.489}$ & $5.490 \pm 1.380$                 \\ \bottomrule
\end{tabular}
\end{table*}

\subsection{Stiff ODEs}

\subsubsection{Stiff ODE in Section~4.2 in the Main Paper}
\citet{ghosh2020steer} propose STEER that shows advantages in learning stiff ODEs, here we apply their experiment to test NDO-NODE. Following their setting, we train all models for $3000$ iterations by RMSprop optimizer with learning rate 0.0001. The training dataset contains 120 points that equally sampled from time range $[0,15]$. We model the state of ODEs as $[x]$, ande use first-order NDO to estimate the derivative $(\dot{x}_i)_{i=1}^N$ from observations $(x_i)_{i=1}^N$. For NDO-NODE and RNODE, we grid search $\lambda$ in range $[10^{-4},1]$. For NDO-NODE, we set $\lambda = 0.05$. For RNODE, we set $\lambda = 0.001$. For NODE with STEER, it uses ODE solver implemented \footnote{See their Github repo at \url{https://github.com/arnabgho/steer}.} by \citet{ghosh2020steer} and the hyperparameter is set to be $b=0.124$.

\textbf{Stiff ODE in Figure 1 in the Main Paper} \citet{ghosh2020steer} also test another stiff ODE
\begin{align}
    \frac{dx}{dt} = -1000x+3000-2000e^{-t} + 1000\sin(t) ~,
\end{align}
with initial condition $x(0)=0$. This stiff ODE is much harder to learn as it vibrates while time $t$ gets larger. It is challenging for NODE to capture both the stiff and fluctuating parts. We train all models for $8000$ iterations, as it is harder to learn. We We grid search $\lambda$ in range $[10^{-4},1]$. For NDO-NODE, we set $\lambda = 0.4$. For RNODE, we set $\lambda = 0.001$. We follow the same setting for STEER and other hyperparameters as \cite{ghosh2020steer}. The results is shown in Figure~1 in the main paper. We can observe that NDO-NODE performs better than others.

\subsection{Airplane Vibration Dataset}
\subsubsection{Experiments in Section~4.2 in the Main Paper}
We follow the setting in \cite{norcliffeSecondOrderBehaviour2020} and test NDO-NODE.
We model the state of the underlying ODE as $[a_2]$. We keep the networks, training, and test process unchanged, and evaluate the first-order NODE. For NDO-NODE, we set $\lambda = 0.005$. For RNODE, we set $\lambda = 0.0005$. 

\subsubsection{Additional Experiments Using Second-order NODE}
As this task aims to learn nonlinear acceleration $a_2$ of the interface on the airplane, \citet{norcliffeSecondOrderBehaviour2020} incorporate physical prior and propose to use their second-order NODE (SONODE) model to try to capture the second-order information. Similar to \cite{norcliffeSecondOrderBehaviour2020}, we test SONODE and NDO-SONODE on this dataset. The neural ODEs in both methods are parameterized as a single hidden layer neural network with 50 hidden units. We train the models on $[0,1000]$ and forecast on $[1000,5000]$ time units. In SONODE, the state of the underlying ODE is defined as $[a_2,z]$, where $z = \frac{\diff a_2}{\diff t}$. Thus, we use both first-order NDO and second-order NDO to estimate the derivatives $[z,\dot{z}]$ and take this estimation as the enhanced signal in NDO-SONODE. We train these models for 1000 iterations by Adam optimizer with an initial learning rate of 0.01 and decayed by a factor of 0.995 at each iteration. For NDO-SONODE, we choose $\lambda = 0.1$.

The results are reported in Figure~\ref{fig:airplane}, which shows NDO-SONODE has a better performance over SONODE on this dataset even SONODE uses physical prior.

\begin{figure*}[!ht]
  \centering
  \includegraphics[width=0.7\textwidth]{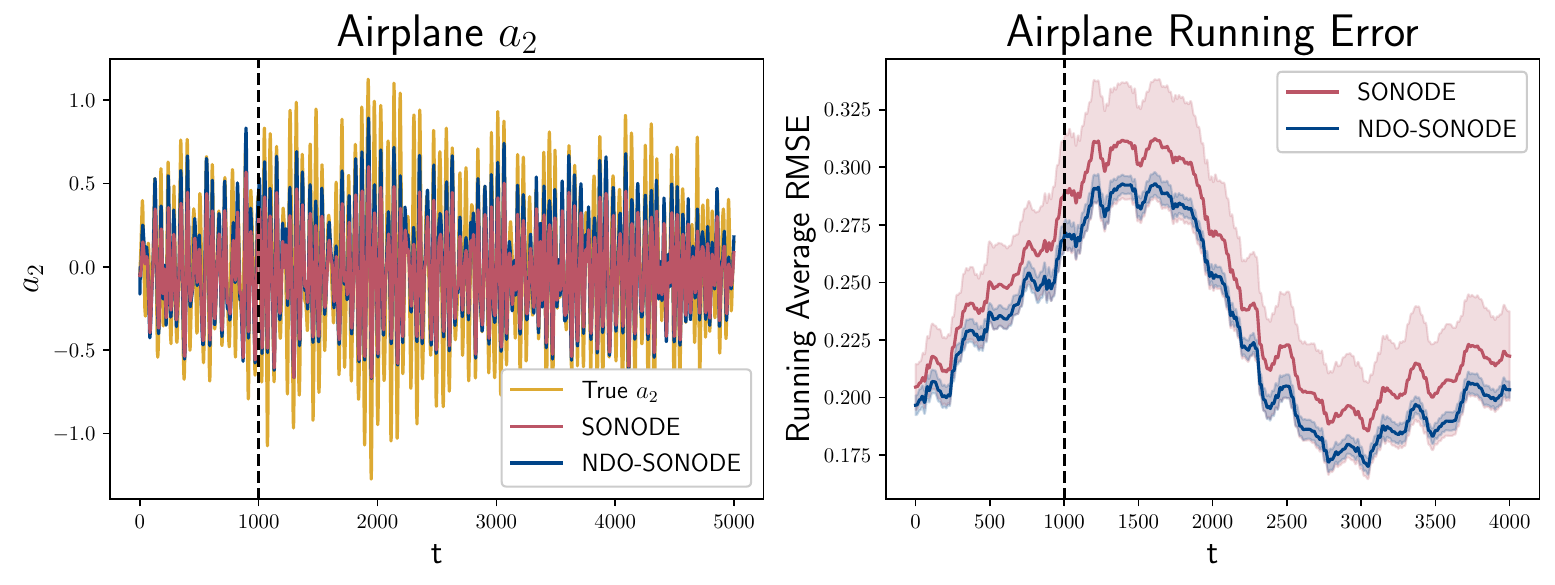}
  \caption{Results of SONODE and NDO-SONODE on airplane vibration dataset. \textbf{(Left)}: ground truth and forecasting trajectories. \textbf{(Right)}: moving averages of root mean square error (RMSE).}
  \label{fig:airplane}
\end{figure*}


\section{Bases for Continuous Function Space}
In our work, we leverage polynomial and trigonometric functions to construct our library as they are bases for the vector space of continuous functions by lemma.\ref{lemma:1} and \ref{lemma:2} from \citet{achieser2013theory}. 
\begin{lemma}[Weierstrass' First Theorem]
Let $f(t)$ be a continuous function on $[a,b]$. Then for any arbitrary $\varepsilon>0$, there exists a polynomial $P(t) = \sum_{i=0}^{\infty} c_i t^i$ on $[a,b]$ such that $\|f-P\|_{\infty} < \varepsilon$. 
\label{lemma:1}
\end{lemma}
\begin{lemma}[Weierstrass' Second Theorem]
Let $f(t)$ be a continuous function on $\mathbb{R}$ that is 1-periodic. Then for any arbitrary $\varepsilon>0$, there exists a trigonometric polynomial $Q(t) = \sum_{i=0}^{\infty} a_i \sin(it) +b_i \cos(it)$ such that $\|f-Q\|_{\infty} < \varepsilon$.
\label{lemma:2}
\end{lemma}

\section{Proof for Theorem 3.1}
\begin{theorem}\label{thm1}Suppose that $\mathcal{Z}_{lib}'\subset\mathcal{Z}_{lib}$ is the training function set for NDO. The Lipschitz constant for the learned neural differential operator function is $L_{NN}$. For given continuous differentiable function $h(t):[T_0,T_1]\rightarrow\mathbb{R}$, we define the distance between two functions as $\rho(h,z)=\frac{1}{N}\sum_{i=1}^N|h(t_i)-z(t_i)|$, where $\{t_i\}_{i=1}^N$ equally partition the time interval $[T_0,T_1]$. $z(t)\in\mathcal{Z}'_{lib}$ is a function in the training data, $h(t)$ is an arbitrary function. The output derivative of NDO for a function is denoted using the subscription $NDO$. Then the error of the output derivation $\dot{h}_{\text{NDO}}$ and the ground truth derivative $\dot{h}$ can be upper bounded as:
{
\small\begin{align*}\displaybreak[3]
\rho({\dot{h}}_{\text{NDO}},\dot{h})
 & \le \frac{L_{NN}}{N}\int_{T_0}^{T_1}|z(t)-h(t)|dt\\
 &+\frac{1}{N}\int_{T_0}^{T_1}|\dot{z}(t)-\dot{h}(t)|dt \\
 &+\frac{|T_1-T_0|^3}{12N^2}M + \rho({\dot{z}_{\text{NDO}}},\dot{z}),
\end{align*}where $M=L_{NN}\cdot\max_{t\in[T_0,T_1]}|\ddot{\epsilon}(t)|+\max_{t\in[T_0,T_1]}|\dddot{\epsilon}(t)|$ with $\epsilon(t)=|z(t)-h(t)|$.}
 \end{theorem}
 \textit{Proof:}
{\small\begin{align*}
\rho&(\dot{h}_{\text{NDO}},\dot{h}) \\
=&\frac{1}{N}\sum_{i=1}^N |\dot{h}_{\textit{NDO}}(t_i)-\dot{h}(t_i)|\\
{=}&\frac{1}{N}\sum_{i=1}^N|\dot{h}_{\textit{NDO}}(t_i)-\dot{z}_{\textit{NDO}}(t_i)+\dot{z}_{\textit{NDO}}(t_i)-\dot{z}(t_i)+\dot{z}(t_i)-\dot{h}(t_i)|\\
{\leq}& \frac{1}{N}\sum_{i=1}^N|\dot{h}_{\textit{NDO}}(t_i)-\dot{z}_{\textit{NDO}}(t_i)|+\frac{1}{N}\sum_{i=1}^N|\dot{z}_{\textit{NDO}}(t_i) -\dot{z}(t_i)|
\\ &+\frac{1}{N}\sum_{i=1}^N|\dot{z}(t_i)-\dot{h}(t_i)|\\
\overset{(1)}{\leq}& L_{\text{NN}}\cdot \frac{1}{N} \sum_{i=1}^N | h(t_i) - z(t_i) | 
+ \frac{1}{N} \sum_{i=1}^N |  \dot{z}_{\text{NDO}}(t_i) -\dot{z}(t_i)| 
\\  & + \frac{1}{N} \sum_{i=1}^N |\dot{z}(t_i) - \dot{h}(t_i) | 
 \\
\overset{(2)}{\leq}& \frac{1}{N}\left(L_{\text{NN}}  \int_{T_0}^{T_1} |h(t) - z(t) |dt+\frac{|T_1-T_0|^3}{12N^2}\max_{t\in[T_0,T_1]}|\ddot{\epsilon}(t)|\right) 
\\
&+\frac{1}{N}\left(\int_{T_0}^{T_1} |\dot{h}(t) -\dot{ z}(t) |dt+\frac{|T_1-T_0|^3}{12N^2}\max_{t\in[T_0,T_1]}|\dddot{\epsilon}(t)|\right)
\\ & + 3\rho(\dot{h}_{\text{NDO}},\dot{h}),
\end{align*}}where 
the inequality $(1)$ is established according to the Lipschitz condition of the neural network \footnote{"The Lipschitz constant for the learned neural differential operator function is $L_{\text{NN}}$" means that $\|\dot{h}_{\text{NDO}}-\dot{z}_{\text{NDO}}\|\leq L_{NN}\|{h}-z\|$, where $\|z\|={\frac{1}{N}\sum_{i=1}^N|z(t_i)|}$.}; the inequality $(2)$ is established according to the numerical error between the integral and its discretization \cite{suli2003introduction} , i.e.,
{\small\begin{align*}
&\big|\sum_{i=1}^N | {h}(t_i) - {z}(t_i)|-\int_{T_0}^{T_1}|h(t)-z(t)|dt\big| \\
\le& \sum_{i=1}^N \big||{h}(t_i) - {z}(t_i) |-\int_{t_i}^{t_{i+1}}|h(t)-z(t)|dt\big|\\
\overset{(3)}{\le}& N\cdot \frac{|T_1-T_0|^3}{12N^2}\max_{t\in[T_0,T_1]}\ddot{\epsilon}(t)\\
\le&\frac{|T_1-T_0|^3}{12N^2}\max_{t\in[T_0,T_1]}\ddot{\epsilon}(t)
\end{align*}}where the inequality $(3)$ is established according to Theorem 7.1 in \cite{suli2003introduction}, and $\epsilon(t)=|h(t)-z(t)|$. Similarly, we have $\big|\sum_{i=1}^N |\dot{h}(t_i) - \dot{z}(t_i)|-\int_{T_0}^{T_1}|\dot{h}(t)-\dot{z}(t)|dt\big| 
\leq\frac{|T_1-T_0|^3}{12N^2}\dddot{\epsilon}(t)$.

\end{document}